\documentclass[journal]{IEEEtran}
\usepackage{graphicx}
\usepackage{booktabs}
\usepackage{caption}
\usepackage{amsmath}
\usepackage{tabularx}
\usepackage{amsfonts}
\usepackage[colorlinks=true, linkcolor=blue, citecolor=blue, urlcolor=blue]{hyperref}
\usepackage{microtype}
\makeatletter
\renewcommand{\fnum@figure}{Fig. \thefigure.\@gobble}
\makeatother
\ifCLASSINFOpdf
\else
\fi
\begin{document}
%
% paper title
% Titles are generally capitalized except for words such as a, an, and, as,
% at, but, by, for, in, nor, of, on, or, the, to and up, which are usually
% not capitalized unless they are the first or last word of the title.
% Linebreaks \\ can be used within to get better formatting as desired.
% Do not put math or special symbols in the title.
\title{Emergency Lane-Change Simulation: A Behavioral Guidance Approach for Risky Scenario Generation}
%
%
% author names and IEEE memberships
% note positions of commas and nonbreaking spaces ( ~ ) LaTeX will not break
% a structure at a ~ so this keeps an author's name from being broken across
% two lines.
% use \thanks{} to gain access to the first footnote area
% a separate \thanks must be used for each paragraph as LaTeX2e's \thanks
% was not built to handle multiple paragraphs
%

\title{Emergency Lane-Change Simulation: A Behavioral Guidance Approach for Risky Scenario Generation}

\author{
Chen~Xiong,
Cheng~Wang,
Yuhang~Liu,
Zirui~Wu,
and~Ye~Tian%
\thanks{Chen Xiong, Cheng Wang, and Yuhang Liu are with the School of Intelligent Systems Engineering, Sun Yat-sen University, Shenzhen 518107, China, and also with the Guangdong Provincial Key Laboratory of Intelligent Transportation System, Sun Yat-sen University, Guangzhou 510275, China.}%
\thanks{Zirui Wu is with the Department of Civil and Environmental Engineering, The Pennsylvania State University, State College, PA 16802, USA.}%
\thanks{Ye Tian is with the School of Traffic Engineering and the Key Laboratory of Road and Traffic Engineering, Ministry of Education, Tongji University, Shanghai 201804, China.}%
\thanks{E-mail: xiongch8@mail.sysu.edu.cn; wangch526@mail2.sysu.edu.cn; liuyh387@mail2.sysu.edu.cn; zrwu@psu.edu; tianye@tongji.edu.cn.}%
\thanks{This work was supported by the Natural Science Foundation of Guangdong Province, China, under Grants 2023A1515012602 and 2025A1515010166, and by the Shenzhen Fundamental Research Program, China, under Grant JCYJ20240813151301003.}%
}

% note the % following the last \IEEEmembership and also \thanks - 
% these prevent an unwanted space from occurring between the last author name
% and the end of the author line. i.e., if you had this:
% 
% \author{....lastname \thanks{...} \thanks{...} }
%                     ^------------^------------^----Do not want these spaces!
%
% a space would be appended to the last name and could cause every name on that
% line to be shifted left slightly. This is one of those "LaTeX things". For
% instance, "\textbf{A} \textbf{B}" will typeset as "A B" not "AB". To get
% "AB" then you have to do: "\textbf{A}\textbf{B}"
% \thanks is no different in this regard, so shield the last } of each \thanks
% that ends a line with a % and do not let a space in before the next \thanks.
% Spaces after \IEEEmembership other than the last one are OK (and needed) as
% you are supposed to have spaces between the names. For what it is worth,
% this is a minor point as most people would not even notice if the said evil
% space somehow managed to creep in.

% The paper headers
\markboth{IEEE Transactions Intelligent Transportation Systems }%
{Shell \MakeLowercase{\textit{et al.}}: Bare Demo of IEEEtran.cls for IEEE Journals}
% The only time the second header will appear is for the odd numbered pages
% after the title page when using the twoside option.
% 
% *** Note that you probably will NOT want to include the author's ***
% *** name in the headers of peer review papers.                   ***
% You can use \ifCLASSOPTIONpeerreview for conditional compilation here if
% you desire.

% If you want to put a publisher's ID mark on the page you can do it like
% this:
%\IEEEpubid{0000--0000/00\$00.00~\copyright~2015 IEEE}
% Remember, if you use this you must call \IEEEpubidadjcol in the second
% column for its text to clear the IEEEpubid mark.

% use for special paper notices
%\IEEEspecialpapernotice{(Invited Paper)}

% make the title area
\maketitle

% As a general rule, do not put math, special symbols or citations
% in the abstract or keywords.
\begin{abstract}
In contemporary autonomous driving testing, virtual simulation has become an important approach due to its efficiency and cost effectiveness. However, existing methods usually rely on reinforcement learning to generate risky scenarios, making it difficult to efficiently learn realistic emergency behaviors. To address this issue, we propose a behavior guided method for generating high risk lane change scenarios. First, a behavior learning module based on an optimized sequence generative adversarial network is developed to learn emergency lane change behaviors from an extracted dataset. This design alleviates the limitations of existing datasets and improves learning from relatively few samples. Then, the opposing vehicle is modeled as an agent, and the road environment together with surrounding vehicles is incorporated into the operating environment. Based on the Recursive Proximal Policy Optimization strategy, the generated trajectories are used to guide the vehicle toward dangerous behaviors for more effective risk scenario exploration. Finally, the reference trajectory is combined with model predictive control as physical constraints to continuously optimize the strategy and ensure physical authenticity. Experimental results show that the proposed method can effectively learn high risk trajectory behaviors from limited data and generate high risk collision scenarios with better efficiency than traditional methods such as grid search and manual design.
\end{abstract}

% Note that keywords are not normally used for peerreview papers.

% For peer review papers, you can put extra information on the cover
% page as needed:
% \ifCLASSOPTIONpeerreview
% \begin{center} \bfseries EDICS Category: 3-BBND \end{center}
% \fi
%
% For peerreview papers, this IEEEtran command inserts a page break and
% creates the second title. It will be ignored for other modes.
\IEEEpeerreviewmaketitle

\section{Introduction}
% The very first letter is a 2 line initial drop letter followed
% by the rest of the first word in caps.
% 
% form to use if the first word consists of a single letter:
% \IEEEPARstart{A}{demo} file is ....
% 
% form to use if you need the single drop letter followed by
% normal text (unknown if ever used by the IEEE):
% \IEEEPARstart{A}{}demo file is ....
% 
% Some journals put the first two words in caps:
% \IEEEPARstart{T}{his demo} file is ....
% 
% Here we have the typical use of a "T" for an initial drop letter
% and "HIS" in caps to complete the first word.
\IEEEPARstart{E}{ffectively} preventing vehicle traffic accidents is a critical issue that must be addressed in the current field of traffic safety research. Among these challenges, emergency lane changes represent one of the most typical high-risk scenarios in traffic accidents, accounting for 23\% of fatal accidents according to data from the National Highway Traffic Safety Administration (NHTSA) \cite{nhtsa2018fars}. Recent studies indicate that autonomous driving technology is widely regarded as a key solution to reducing traffic accidents, with its core advantage lying in minimizing human error through advanced algorithms and sensors \cite{abdel2024matched}. However, ensuring the safety of self-driving vehicles remains a significant challenge before their deployment across diverse road environments. Traditional real-world road testing faces limitations such as lengthy test cycles, high costs, and insufficient scene diversity. As noted in relevant research \cite{kalra2016driving}, Levels 3 and above in autonomous driving require safety evaluations spanning hundreds of millions of kilometers to ensure reliability. In contrast, virtual simulation-based testing offers rich scenario content, higher efficiency, and lower costs \cite{zhong2021survey,fremont2020formal}, demonstrating considerable potential. This method is increasingly considered an essential component of the evaluation framework for autonomous vehicles \cite{cai2022survey,tang2023survey,zhang2023novel}. \par

Data collection for virtual scene testing primarily depends on natural driving datasets (NDD) and high-test-value scenario data. To date, Waymo has performed over 32 billion kilometers of autonomous driving tests in virtual environments \cite{team2023self}, yet it remains unable to provide comprehensive solutions for autonomous driving across diverse long-tail scenarios. Previous research has highlighted that high-value test scenarios will become a bottleneck in the development of autonomous driving \cite{killing2021learning}. Constructing high-value test scenarios is a challenging task, with collision scenarios being particularly critical due to their high accident costs. In mixed traffic flow environments, most collision scenarios result from human driver operational errors \cite{zhou2024would, schwall2020waymo}. However, collision scenarios collected from real-world data are low-probability events, and the limited availability of such data leads to scenario constraints, making it difficult to develop a robust autonomous driving solution based on a small number of collision cases.  \par

Among existing methods for generating risky traffic scenarios, two critical challenges persist: effectively learning emergency vehicle behaviors from limited data, and generating scenarios that adhere to the physical constraints of vehicle dynamics. In the domain of vehicle trajectory learning, recurrent neural architectures such as Recurrent Neural Networks (RNNs) and Long Short-Term Memory (LSTM) networks have been widely adopted due to their ability to capture temporal dependencies and simulate complex nonlinear dynamics \cite{katariya2022deeptrack, hui2022deep, xie2024advdiffuser}. However, these methods often struggle to extract representative features of dangerous trajectories from sparse data and suffer from limited generalization capabilities caused by model convergence tendencies. In contrast, Generative Adversarial Networks (GANs) offer a promising alternative by leveraging adversarial training to generate diverse trajectory patterns. GAN-based approaches are particularly effective in unsupervised or semi-supervised settings, allowing for efficient utilization of limited data and reducing the reliance on extensive data annotation \cite{demetriou2023deep, jia2023conditional, yu2017seqgan}. On another front, current research often focuses on mining special traffic scenarios from large-scale Naturalistic Driving Data (NDD) and subsequently using models such as diffusion networks to guide vehicle trajectories toward targeted objectives, thereby constructing adversarial scenarios by combining potentially risky behaviors \cite{feng2021intelligent, feng2023dense, xu2023diffscene, hao2023bridging}. Despite their effectiveness, these approaches generally rely on extensive trajectory extraction from NDD and lack mechanisms to directly incorporate the physical feasibility of vehicle movements. As a result, their strong dependence on large datasets and limited consideration of physical constraints pose significant challenges. Moreover, their capability to generate realistic and physically plausible risky scenarios under data-scarce conditions remains questionable. \par

\begin{figure}
	\centering 
	\includegraphics[width=0.48\textwidth]{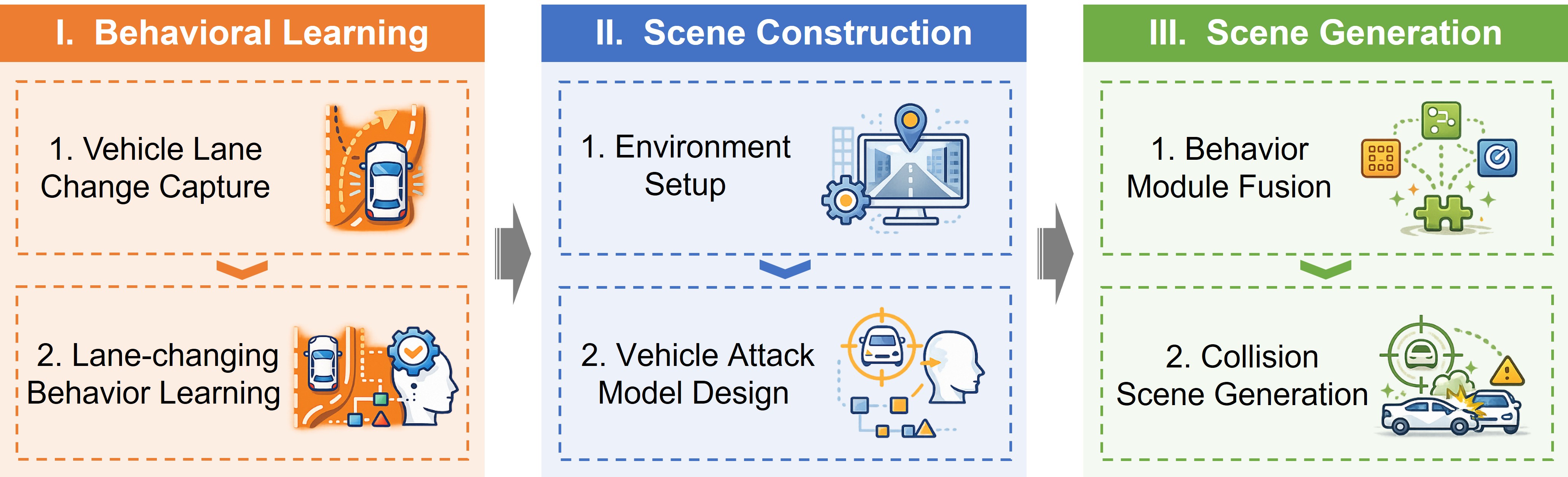}	
	\caption{\textcolor{red}{Overall Model Framework}} 
	\label{1}%
\end{figure}

 In response to the aforementioned challenges, this study aims to guide vehicles in generating risky scenarios through the learning of emergency lane-changing behaviors. To establish a reliable benchmark for subsequent model training, we first extract representative lane-changing trajectories of emergency vehicles from the HighD dataset (Highway Drone Dataset) \cite{krajewski2018highd}, ensuring the behavioral authenticity and fidelity of the data. Then, to address the scarcity of high-quality emergency lane-change data, we enhance the Sequence Generative Adversarial Network (SeqGAN) \cite{yu2017seqgan} by improving its adversarial training strategy. This modification enables the model to better capture the spatiotemporal dynamics of high-risk lane-changing maneuvers. Furthermore, to simulate realistic and aggressive driving behaviors under hazardous conditions, we adopt the Recurrent Proximal Policy Optimization (RecurrentPPO) algorithm \cite{pleines2023memory}. Leveraging its recurrent architecture, the model effectively encodes temporal dependencies in driving decisions, facilitating the generation of high-risk trajectories that closely resemble real-world behavior. To ensure physical feasibility in simulated risky scenarios, the trajectories generated by RecurrentPPO are further refined using Model Predictive Control (MPC) and validated interactively in the CARLA simulator \cite{dosovitskiy2017carla}, where MPC imposes physical constraints to guarantee scenario plausibility. Lastly, to enhance the accuracy of hazardous scenario recognition, we optimize the time-to-collision (TTC) computation and improve environmental perception mechanisms, thereby strengthening the system’s capability in risk identification and prediction. The framework proposed in this study leverages behavior learning to enable the attack vehicle to conduct active attacks, thereby creating realistic interactive risky test scenarios, as illustrated in Figure \ref{1}.

The contributions of this study are summarized as follows:\par

\begin{itemize}

\item We propose an optimized Sequence Generative Adversarial Network (GA-SeqGAN) to enhance the learning and data generation capabilities for few-shot emergency lane-changing trajectory behaviors.

\item A behavior learning module is introduced and combined with a novel active attack strategy, enabling the ego vehicle to perform emergency lane changes based on the learned behaviors. The generated behaviors closely align with the characteristics of real-world trajectory data.

\item Furthermore, physical constraints of the vehicle are incorporated into the control framework. The ego vehicle executes the learned attack strategies and interacts with background vehicles within a simulation platform, ensuring the physical feasibility of the generated scenarios. This integration effectively improves the system’s ability to perceive surrounding risks and facilitates the generation of realistic and risky traffic scenarios.

\end{itemize}

 Section 2 provides a comprehensive review of the current research landscape and identifies the existing research gaps. Section 3 details the procedure for generating risky collision scenarios during emergency lane changes, incorporating the combined behavioral guidance module. Section 4 assesses the fidelity of the behavior simulation, evaluates the reasonableness of the reward function design, and discusses the efficacy of our proposed method based on the simulation outcomes.\par

\section{Relate Work}
\subsection{Virtual Test Scenario Generation}
Unlike traditional scenario generation methods, the scenario-based virtual testing shifts autonomous driving safety evaluation from the real world to a virtual environment, enhancing vehicle adaptability and reducing the need for frequent adjustments as autonomous vehicle systems evolve. The generation of virtual test scenarios is a critical component of autonomous driving testing. Relevant Scholars have categorized scenario generation methods into three main approaches: knowledge-based generation, data-driven generation, and adversarial generation \cite{ding2023survey, pronovost2023scenario, zhang2024chatscene, heuer2022scenario, rafiei2022pedestrian, tan2021scenegen}. However, these methods have limitations in adapting to potential critical scenarios necessary for addressing emerging challenges in autonomous driving. These challenges primarily involve the response capabilities to dynamic and complex interactions as well as the constantly evolving spatiotemporal characteristics of the driving environment. \par
Deep learning and reinforcement learning are effective solutions in existing research. In the context of dynamic and complex interactions, Li et al. \cite{li2024autonomous} apply game theory to autonomous driving decision-making models for multi-vehicle interactions. Jenkins et al. \cite{jenkins2018accident} develop a method that uses recurrent neural networks to generate test scenarios by inducing artificial crash situations. Regarding the ability to respond to spatiotemporal characteristics and internal elements of dynamic scenarios, Hoel et al. \cite{hoel2019combining} introduce a tactical decision-making framework that employs reinforcement learning for decision-making. Lu et al. \cite{lu2022learning} develop Deep Collision, in which collision probability serves as a safety metric used to define the reward function, enabling the system to learn the environmental conditions leading to autonomous vehicle collisions. Reinforcement learning based on signal temporal logic is applied to generate critical vehicle interaction scenarios, including following and cut-in driving scenarios \cite{qin2019automatic}.While these studies effectively explore critical scenarios, whether the generated vehicle behavior trajectories align with human driving behavior remains uncertain.

\subsection{Vehicle Trajectory Behavior Learning}
To learn driving behavior and generate realistic vehicle trajectories, Song et al. \cite{song2022learning} utilize a model-based generator to produce future trajectories under explicit constraints, combining it with an evaluator to select the most appropriate trajectory for vehicle trajectory prediction. Lu captures historical vehicle trajectories and constructs a context-aware graph convolutional network with temporal attention to predict vehicle intentions \cite{lu2022vehicle}. To capture interactions among surrounding vehicles, Zuo et al. \cite{zuo2023trajectory} develop a novel trajectory prediction network based on a multi-head attention mechanism. While these methods effectively learn vehicle behavioral intentions, they struggle to autonomously generate novel trajectories with similar distributions.\par
An effective approach to learning vehicle behavior while generating trajectories that conform to the dataset distribution is the application of adversarial neural networks for adversarial learning. Demetriou et al. \cite{rossi2021vehicle} employ generative adversarial networks (GANs) to generate trajectories of varying lengths based on real-world data. Jia et al. \cite{jia2023conditional} use CTGAN to capture motion patterns and provide feedback based on the degree of alignment with latent intentions, thereby generating multiple candidate future trajectories.Among these methods, SeqGAN \cite{yu2017seqgan} demonstrates superior performance in learning sequential information and holds significant potential for trajectory behavior learning. Han et al. \cite{han2021variable} leverage an improved feature-generating adversarial network to predict complex dynamic behaviors. Jing et al. \cite{jing2024efficient} apply a variant of SeqGAN to extract emergency lane-change trajectories from datasets, effectively capturing human driving behavior. However, solely relying on generative adversarial networks is insufficient for guiding vehicle behavior and controlling vehicles to exhibit hazardous actions based on the learned information

\subsection{Vehicle Active Conflict Strategy}
Deep reinforcement learning (DRL) effectively integrates trajectory sequence information, enabling it to address challenges related to vehicle control dynamics, temporal dependencies, and partial observability \cite{hui2022deep}. Chen et al. \cite{chen2021adversarial} propose an adaptive evaluation framework for creating challenging scenarios, leveraging adversarial environments generated by DRL to conduct dynamic testing and evaluation of autonomous vehicles. Siboo et al. compare the PPO and DDPG algorithms, concluding that DDPG is better suited for autonomous vehicle control \cite{siboo2023empirical}. In terms of practical applications, Acquarone finds that the TD3 algorithm is particularly effective in handling challenges involving continuous states and actions, making it well-suited for vehicle motion control \cite{acquarone2023cooperative}. However, these algorithmic strategies still face difficulties in integrating sequential trajectories with strong temporal dependencies and adapting to complex dynamic scenarios. In response to this challenge, Trauth et al. \cite{trauth2024reinforcement} employ the RPPO method to develop autonomous driving systems capable of effectively managing complex situations and adapting to evolving environments. \par
Existing studies primarily rely on hazardous scenario data extracted from naturalistic driving datasets (NDDs), yet the limited data volume imposes constraints on the learned behaviors. Additionally, during scenario generation, whether the attack vehicle’s trajectory aligns with human driving behavior is often overlooked. To address these research gaps, we design a behavior learning module to generate realistic human driving trajectories and guide a vehicle attack module for controlling the attack vehicle. The controlled attack vehicle reacts to the autonomous vehicle’s actions during its maneuvering process, simulating high-risk interactions and generating hazardous collision scenarios.

\section{Methodology}%the citaion havn't mention
In this study, we propose a framework for generating high-value test scenario data, which consists of three main stages: (1) learning vehicle emergency lane-change behaviors, (2) designing an active attack strategy for the ego vehicle, and (3) generating emergency lane-change test scenarios. The emergency lane-change behavior is generated by the GA-SeqGAN based on the HighD dataset. Combined with the previous literature review, it is evident that relatively limited research has been conducted on dangerous behaviors in typical urban road scenarios. Consequently, this study aims to focus on generating risky scenarios for emergency lane-changing behaviors on urban expressways, providing a detailed elaboration of the technical aspects involved. 

\subsection{Data Processing}
To obtain and generate human-driven emergency lane-change behaviors, this study utilizes the publicly available German HighD dataset \cite{krajewski2018highd}. This dataset comprises vehicle driving data collected from six different highway locations, covering over 110,500 vehicles recorded over 11.5 hours. The dataset includes vehicle positions, instantaneous velocities, instantaneous accelerations, and other vehicle-related data recorded at a time interval of 0.025 seconds.We adopt the emergency lane-change behavior identification method proposed by Jia et al. \cite{zhang2023recognition} to identify and extract emergency lane-change behaviors from the dataset. Based on the extracted emergency lane-change data, we select lateral position change ($X$) and longitudinal velocity change ($V$) as key parameters of the lane-change behavior \cite{zhao2023generalization, jing2024efficient}. In this study, emergency lane-change data with a lane-change duration of less than 2 seconds are selected.To ensure clarity and simplicity of the trajectory data, we normalize the coordinates of the lane-change trajectories and set the starting coordinates of each trajectory to zero. Using the extracted data, we construct a dataset comprising 520 emergency lane-change trajectories.

\subsection{Behavior Learning Module}
Current datasets are typically collected from naturalistic driving, where the number of critical lane-change interaction scenarios is extremely limited and insufficient to cover various emergency lane-change behaviors. To address this limitation, this study introduces SeqGAN, a technique capable of effectively extracting sequence data features, to expand the dataset. GANs have been widely applied in text generation, image processing, and computer vision \cite{demetriou2023deep, chen2021adversarial, gan2022higan+, wang2021generative}.A GAN consists of two neural networks: a generator and a discriminator. 

The generator takes random noise as input and maps it to actual data \( x \) through the function \( G(z, \theta_g) \), where \( \theta_g \) denotes the parameters of the generator neural network. The discriminator aims to determine whether the input data is the original data \( x \) or synthetic data generated by the generator, and assigns appropriate labels accordingly. By training \( D(x, \theta_d) \), the discriminator progressively enhances its capability to accurately differentiate between real and synthetic data, where \( \theta_d \) represents the parameters of the discriminator. Simultaneously, the generator's objective is to produce data that can deceive the discriminator into classifying it as genuine data. These two neural networks engage in a competitive relationship, mutually driving each other to optimize their internal parameters and generate data indistinguishable from real data via adversarial training.
At the same time, the generator aims to produce data that can deceive the discriminator into misclassifying synthetic data as real. This adversarial relationship drives both networks to optimize their internal parameters iteratively. Through adversarial training, the generator produces data that becomes increasingly indistinguishable from real data.
The adversarial outcome between $G$ and $D$ is represented by the value function $V(D,G)$, as shown in Equation (1):
\begin{equation}
\begin{aligned}
\min_{G} \max_{D} V(D, G) &= \mathbb{E}_{z \sim p_z(z)} \left[ \ln(1 - D(G(z))) \right] \\
&\quad + \mathbb{E}_{x \sim p_{\text{data}}(x)} \left[ \ln D(x) \right]
\end{aligned}
\end{equation}
where \( p_z(z) \) represents the prior distribution of the input noise variable. Under ideal conditions, the loss function in Equation~(1) is equivalent to the Jensen-Shannon (JS) divergence.
\begin{equation}
\begin{aligned}
JSD\left(p_{data}\parallel p_g\right) &= \frac{1}{2}KL\left(p_{data}\parallel\frac{p_{data}+p_g}{2}\right) \\
&\quad + \frac{1}{2}KL\left(p_g\parallel\frac{p_{data}+p_g}{2}\right)
\end{aligned}
\end{equation}
where \( p_g \) represents the distribution of samples \( z \sim p_z \) in \( G(z) \). \par
The Kullback-Leibler (KL) divergence, as a maximum likelihood estimation method, effectively quantifies the relationship between two distributions. If there is no overlap between the two distributions, the KL divergence approaches infinity; if the two distributions are identical, the KL divergence converges to 0, yielding the unique solution \( p_g = p_{\text{data}} \). The formula for calculating KL divergence is: 
\begin{equation}
KL\left(p_{\text{data}} \parallel p_g\right)
= \sum_i p_{\text{data}}(i)\log\left(\frac{p_{\text{data}}(i)}{p_g(i)}\right)
\end{equation}
Compared to traditional GANs, SeqGAN exhibits significant advantages in handling continuous and time-series data. By employing a framework that integrates a generator and a discriminator, SeqGAN effectively models sequential dependencies, enabling the generation of continuous data with temporal correlations. To address the imbalance between generated and real data, Monte Carlo (MC) search is utilized to estimate the expected reward for a complete sequence generated by the generator. This approach ensures that the generated data distribution more closely approximates the real data distribution, as formulated in the following equation.
\begin{equation}
Q_\theta\left(y_{1:t},a_t\right) = \mathbb{E}_{y_{t+1:T} \sim G_\theta} \left[ R\left(y_{1:T}\right) \mid y_{1:t} \right]
\end{equation}
where \( Q_\theta(y_{1:t}, a_t) \) represents the expected reward after taking action \( a_t \) at time step \( t \) under policy \( G_\theta \). \( R(y_{1:T}) \) denotes the final reward of the complete sequence \( y_{1:T} \). \( y_{1:t} \) refers to the sequence generated up to time step \( t \), while \( y_{t+1:T} \sim G_\theta \) represents the remaining sequence sampled from the generator \( G_\theta \). \par
To enhance SeqGAN’s ability to learn short sequences, we introduce an optimization to the original framework. The original SeqGAN employs a convolutional neural network (CNN) as the backbone of its discriminator. However, CNNs are less effective than gated recurrent units (GRUs) in handling sequential information and variable-length sequences. Specifically, CNNs rely on local receptive fields in their convolutional operations, making it difficult to capture long-range temporal dependencies. In contrast, GRUs leverage their gating mechanisms, including update and reset gates, to model long-term dependencies in sequential data more effectively.Additionally, CNNs contain a large number of parameters, leading to extended training times and an increased risk of overfitting. To improve the learning capability for short sequences while reducing the model complexity, we replace the CNN-based discriminator with a GRU-based structure. The GRU formulation enables GA-SeqGAN to better capture temporal dependencies, process continuous variables, and generate dynamic sequences. The parameter update equations for GRU are given as follows:
\begin{equation}
z_t=\sigma\left(W_z\cdot\left[h_{t-1},x_t\right]\right)
\end{equation}
\begin{equation}
r_t=\sigma\left(W_r\cdot\left[h_{t-1},x_t\right]\right)
\end{equation}
\begin{equation}
\widetilde{h_t}=\tanh\left(W\cdot[r_t\odot h_{t-1},x_t]\right)
\end{equation}
\begin{equation}
h_t=\left(1-z_t\right)\odot h_{t-1}+z_t\odot\widetilde{h_t}
\end{equation}
where \( z_t \) and \( r_t \) represent the update gate and reset gate, respectively, while \( h_{t-1} \) denotes the hidden state at the previous time step, and \( x_t \) is the input at the current time step. The function \( \sigma \) refers to the sigmoid activation function, and \( \bigodot \) denotes element-wise multiplication. By applying this transformation, the discriminator better captures the temporal dependencies in the data, handles continuous variables, and generates dynamic sequences. \par
Meanwhile, to ensure that the generator and discriminator possess equivalent capabilities, optimizations are also applied to the generator. The original SeqGAN generator consists of an input layer, an embedding layer, a recurrent neural network (RNN), and an output layer. To enhance the generator's ability to model sequential data, an attention mechanism (\texttt{AttentionLayer}) is introduced after the embedding layer. The attention mechanism is formulated as follows:
\begin{equation}
e_{t,i}=v^Ttanh\left(W_hh_t+W_ss_i+b\right)
\end{equation}
\begin{equation}
\alpha_{t,i}=\frac{exp\left(e_{t,i}\right)}{\sum_{j=1}^{T}exp\left(e_{t,j}\right)}
\end{equation}
\begin{equation}
c_t=\sum_{i=1}^{T}\alpha_{t,i}s_i
\end{equation}
where \( h_t \) represents the hidden state at the current time step, \( s_i \) denotes the representation of the \( i \)-th element in the sequence, and \( v \), \( W_h \), \( W_s \), and \( b \) are learnable parameters. \( \alpha_{t,i} \) signifies the attention weight, while \( \mathbf{c}_t \) is the context vector. By incorporating the attention mechanism, the generator effectively focuses on critical parts of the sequence, thereby enhancing the quality of generated sequences.\par

Compared to LSTM, GRU features a more streamlined structure with fewer parameters and faster training, while maintaining comparable or superior performance in certain tasks. Replacing LSTM with GRU in the generator accelerates convergence and enhances training stability. Through joint optimization of the generator and discriminator, SeqGAN achieves improved temporal modeling for short sequences, reduces parameter complexity, and enhances training efficiency and sequence quality. \par

\begin{figure}
	\centering 
	\includegraphics[width=0.38\textwidth]{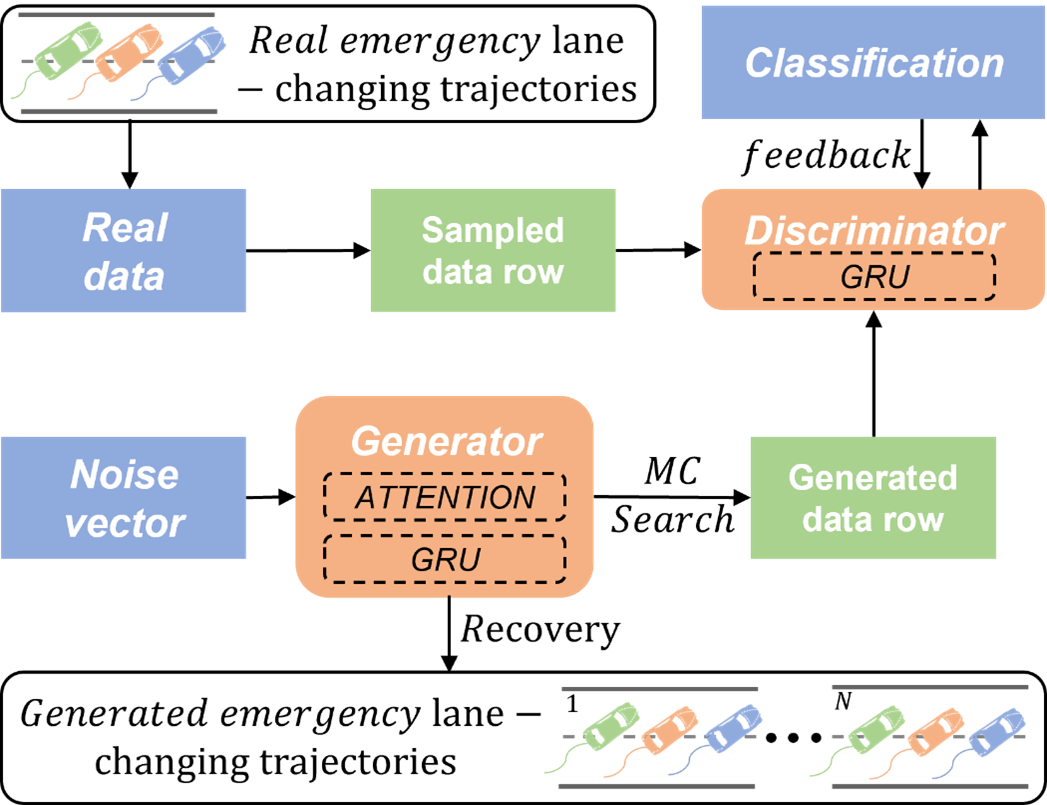}	
	\caption{The GA-SeqGAN Training Framework} 
	\label{2}%
\end{figure}

Additionally, to maintain balance between the generator and discriminator, the generator architecture is further optimized. The original SeqGAN generator comprises an input layer, an embedding layer, a recurrent neural network (RNN) core, and an output layer. To constrain data variations within a controllable range, an attention layer is introduced following the embedding layer. Furthermore, the RNN core is updated from LSTM to GRU, facilitating faster convergence and enhanced training stability. The detailed structure of GA-SeqGAN is illustrated in  Figure \ref{2}. After training, the generator produces realistic and diverse dangerous 

\subsection{Vehicle Active Attack Strategy}

After learning the emergency lane-changing behavior of vehicles under manual driving conditions, embedding an active attack driving strategy into background vehicles becomes essential. This approach transforms high-risk scenarios from passive occurrences into actively triggered events. At the implementation level, constructing an appropriate algorithm is necessary to generate the future trajectory reference sequence for the attacking vehicle. Since vehicles rely only on local information about surrounding traffic during decision-making and exhibit strong temporal dependencies, RPPO is adopted to facilitate vehicle behavior decision-making and state optimization. Building on this foundation, integrating the vehicle behavior guidance module with the RPPO-based active attack strategy leads to the development of the Vehicle Active Attack Model (VAA).\par

RPPO is a reinforcement learning algorithm designed for partially observable environments. It utilizes recurrent neural networks (e.g., LSTM or GRU) to model temporal correlations, enabling efficient decision optimization in dynamic and complex scenarios. This algorithm is well-suited for path planning and behavior control tasks in autonomous driving systems, maximizing cumulative rewards through trial-and-error learning and allowing vehicles to navigate complex environments according to predefined goals. The key strength of RPPO lies in its ability to capture dynamic interaction patterns by integrating time-series information while improving vehicle control stability through policy optimization. Based on Partially Observable Markov Decision Processes (POMDP), RPPO defines the state space \( S \), action space \( A \), state transition probability \( T \), reward function \( R \), observation space \( \Omega \), and observation model \( O \). To address the limitations of POMDP, RPPO employs recurrent neural networks to model state sequences, thereby enhancing policy performance in partially observable environments.\par

In scenario design, RPPO extracts dynamic features from vehicle observations, including position, speed, acceleration, and orientation angle. Additionally, sensor data captures other dynamic environmental factors, such as background vehicles and obstacles, which serve as state inputs for modeling. After normalizing these observations, the GRU module generates time-series features to construct a deep representation of the current state. The policy optimization process in RPPO is based on the clipped objective function of PPO, which stabilizes learning by preventing excessive policy updates. The objective function is:
\begin{equation}
L^{\text{CLIP}}(\theta) = \mathbb{E}_t \left[ \min \left( r_t(\theta) A_t, \text{clip} \left( r_t(\theta), 1-\epsilon, 1+\epsilon \right) A_t \right) \right]
\end{equation}

Here, \( r_t(\theta) = \frac{\pi_\theta(a_t \mid s_t)}{\pi_{\theta_{\text{old}}}(a_t \mid s_t)} \), where \( \pi_\theta \) denotes the policy network, \( A_t \) represents the advantage function, and \( \epsilon \) is the clipping threshold. The advantage function \( A_t \) is computed based on the temporal difference of value functions, ensuring that the policy update remains consistent with the actual state estimation. The formula is presented as follows:
\begin{equation}
A_t = r_t + \gamma V(s_{t+1}^{\text{ppo}}) - V(s_t^{\text{ppo}})
\end{equation}
where, \( V(s_t^{\text{ppo}}) \) is the default value function, and \( \gamma \) is the discount factor. \par
In RPPO, recurrent neural networks (e.g., LSTM or GRU) are employed to model time-series information, and their parameters are optimized via backpropagation. During each gradient update, the parameter adjustment formula is as follows:
\begin{equation}
\theta\gets\theta-\alpha\nabla_\theta\ L^CLIP\ (\theta)
\end{equation}
where, \(\alpha\) denotes the learning rate, and \(\nabla_\theta L^{CLIP}(\theta)\) represents the gradient of the loss function with respect to the model parameters.\par
To realize the emergency lane-changing attack behavior of a vehicle against surrounding vehicles, the reward function must be carefully designed. To ensure that the vehicle is incentivized to exhibit attack behaviors toward surrounding vehicles, parameters such as inter-vehicle distance, acceleration changes, and collision occurrence are incorporated. At the same time, to ensure that the emergency lane-changing maneuver aligns with human driving behavior, cosine similarity is employed as a metric to evaluate the consistency between the vehicle's actual trajectory \(f(x_i)\) and the generated trajectory \(g(x_i)\). The formula is presented as follows:
\begin{equation}
\mathrm{cos\_sim}
=
\frac{\sum_{i=1}^{n} f(x_i)\,g(x_i)}
{\sqrt{\sum_{i=1}^{n} f(x_i)^2}\,\sqrt{\sum_{i=1}^{n} g(x_i)^2}}
\end{equation}
where, l denotes the minimum deviation distance between the main vehicle and the target path; lane indicates the lane deviation status; \(\nabla a\) represents the change in acceleration; \(\mathrm{cos_sim}\) measures the deviation of the learned trajectory from the target trajectory; collision signifies the angular difference between the direction of the main vehicle and the target direction; collision\ reflects the collision status of the vehicle; and baseline serves as a reference to enable more comprehensive strategy exploration. Additionally, \(\lambda_1,\lambda_2,\lambda_3,,\lambda_4,\lambda_5\) are weight factors assigned to each term. By dynamically adjusting these weights, the reward function can be adaptively optimized.\par
\begin{figure}
	\centering 
	\includegraphics[width=0.43\textwidth]{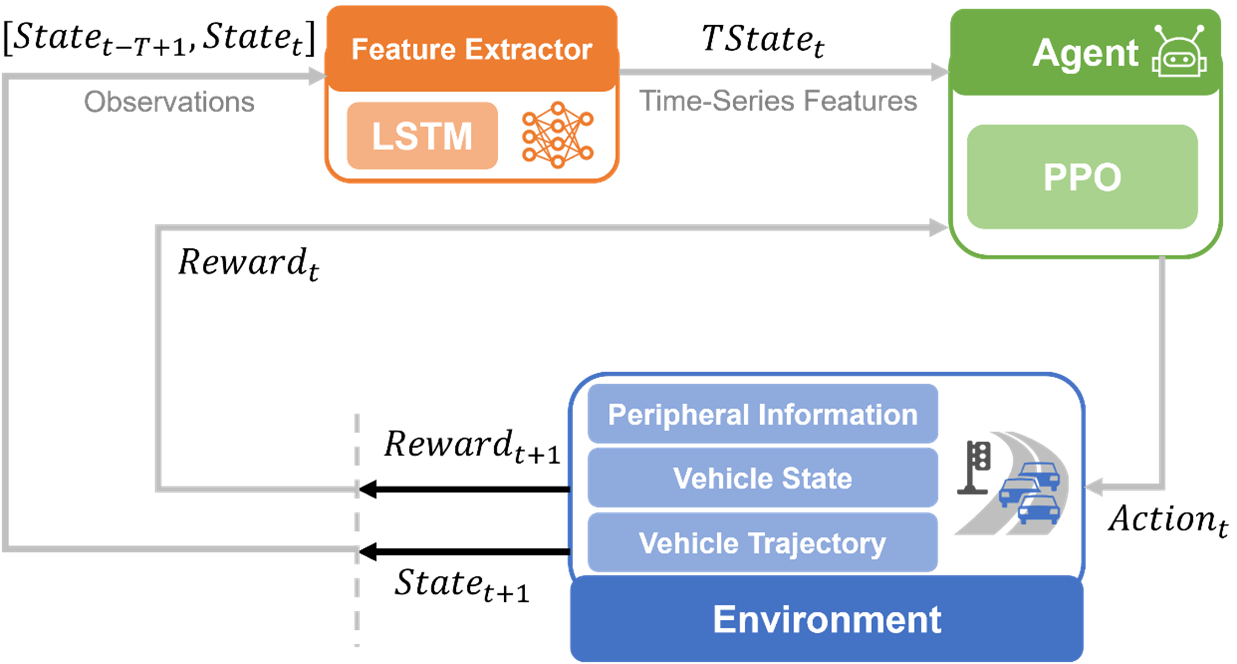}	
	\caption{The VAA Framework} 
	\label{3}%
\end{figure}

During the training process, RPPO employs an experience replay buffer to store state-action sequences and jointly optimizes the policy network and value function network via segment-based sampling. As illustrated in  Figure \ref{3}, after each sampling step, the recurrent neural network resets its hidden state across time steps to ensure the continuity and stability of the training process.\par
\subsection{Method for Generating Risky Lane-Change Scenarios in Vehicle Dynamics}
The vehicle active attack model is capable of generating the future trajectory sequence of a vehicle; however, it does not facilitate genuine physical interactions. Furthermore, the dynamics of the traffic environment and other physical factors must be considered and integrated into the overall framework. To achieve this, during the design of the physical scene, we input the state information of the vehicle into the Vehicle Active Attack (VAA) model to generate a reference trajectory sequence for the attacking vehicle. Subsequently, these trajectories are utilized as control commands for a Model Predictive Controller (MPC) based on the Carla platform \cite{dosovitskiy2017carla}. The MPC ensures precise vehicle control while providing feedback to the optimizer, enabling continuous refinement of attack strategies and vehicle behavior decisions.\par
\begin{figure}
	\centering 
	\includegraphics[width=0.48\textwidth]{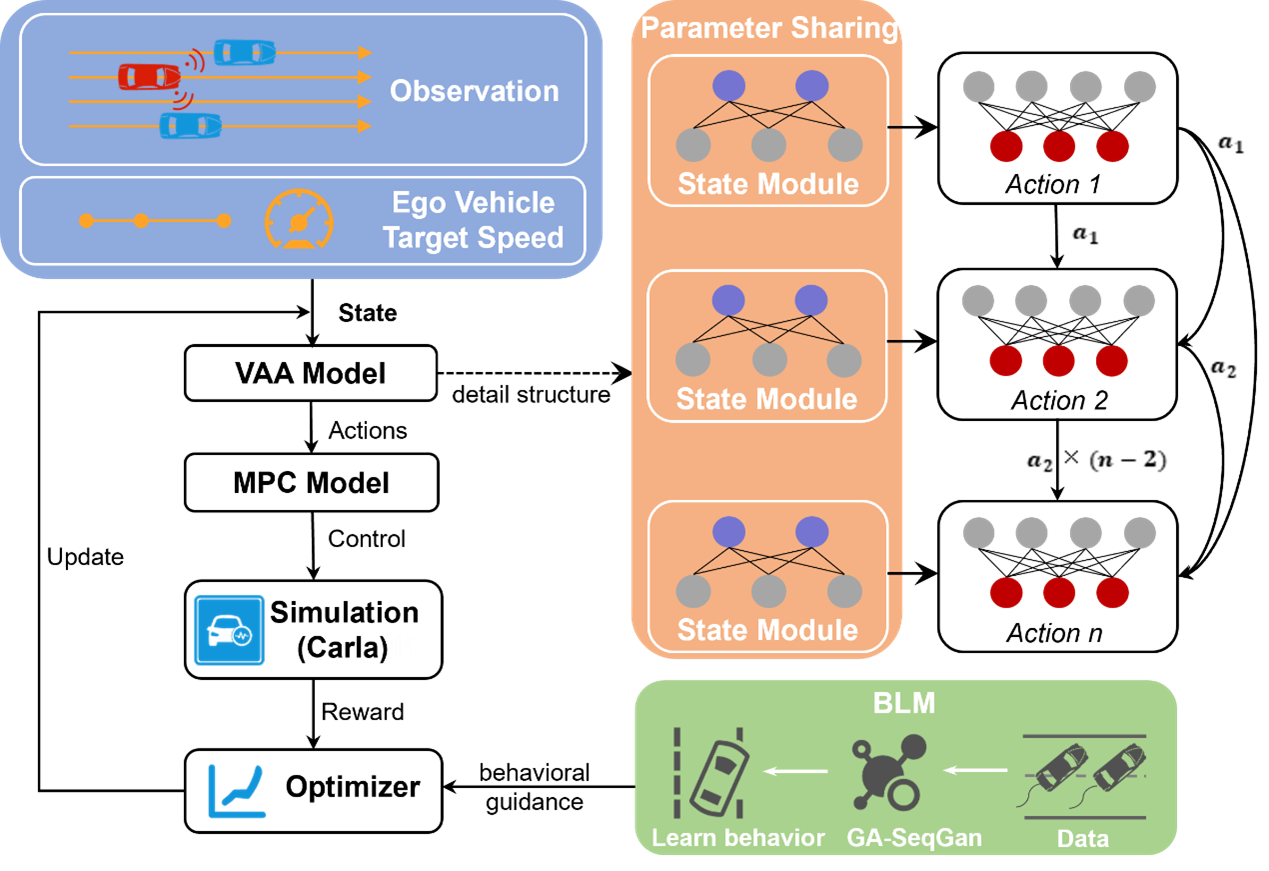}	
	\caption{Method for Generating Risky Lane-Change Scenarios in Vehicle Dynamics} 
	\label{4}%
\end{figure}

Through this approach, real physical interactions are realized, and the variability of complex traffic environments is accounted for. This ensures that the background vehicle dynamically responds to environmental changes in attack scenarios and effectively interacts with other vehicles and adheres to traffic rules. Throughout this process, the closed-loop feedback mechanism between the VAA model and the MPC controller provides real-time feedback for strategy training, allowing the attacking vehicle to continuously adjust and optimize its behavior strategy within the real physical environment, as illustrated in Figure \ref{4}.\par

In the experiment, when the active attack vehicle solely relies on deep reinforcement learning to explore dangerous scenarios, issues such as poor convergence and susceptibility to local optima arise. To enable the active attack vehicle to achieve collision more efficiently, we integrated Time-to-Collision (TTC) detection into its decision-making process. The formula for TTC is provided below:
\begin{equation}
\begin{aligned}
TTC &= \min \left( TTC_1, TTC_2, \ldots, TTC_j \right) = \\
&\begin{cases} 
\dfrac{\sqrt{\left( x_{j} - x_{E} \right)^2 + \left( y_{j} - y_{E} \right)^2}}{
\nabla v \cdot \cos \left( \arctan \left( \dfrac{y_{j} - y_{E}}{x_{j} - x_{E}} \right) - \theta_{E} \right)} & \text{ if } \nabla v > 0\\
\infty & \text{otherwise}
\end{cases}
\end{aligned}
\end{equation}
where \(\nabla v\) represents the relative velocity of the vehicle. \(x_\mathrm{j},x_{\mathrm{E}},y_\mathrm{j},y_{\mathrm{E}} \) respectively denote the horizontal and vertical coordinate positions of background vehicle \(j(j\in J)\) and the designed vehicle \(\partial_{\mathrm{E}}\) indicates the heading angle of the Ego vehicle.\par

\begin{figure}
	\centering 
	\includegraphics[width=0.48\textwidth]{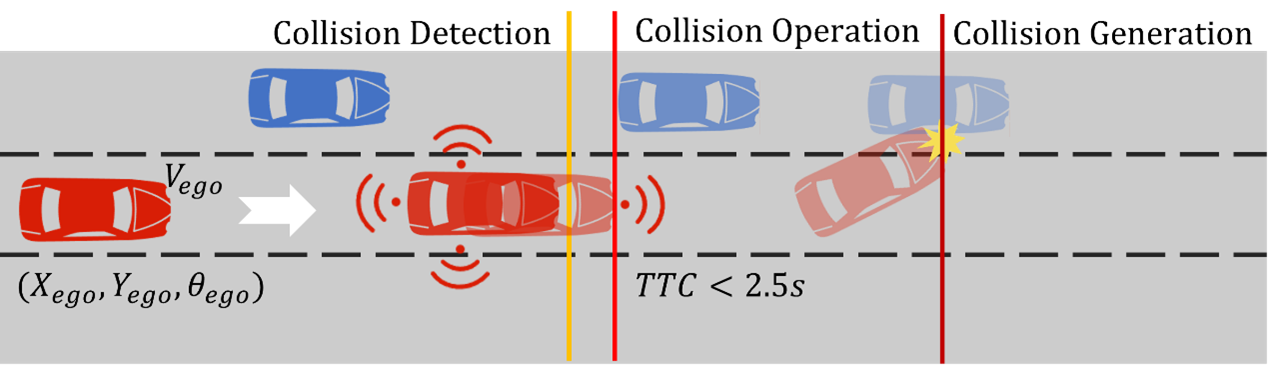}	
	\caption{The Triggering Process of the Emergency Lane Change Strategy  } 
	\label{5}%
\end{figure}

Scenarios that are difficult for humans to handle within their reaction time should be addressed by autonomous driving systems. In this context, when the Time-To-Collision (TTC) is less than the human reaction threshold of 2.5 seconds \cite{xu2021robust}, the designed vehicle is triggered to initiate an attack strategy, as illustrated in Figure \ref{5}. In the generated scenarios, the RecurrentPPO algorithm first generates an optimal action \( A_i \), which is then used to compute a set of reference waypoints based on the current state and position. These waypoints are subsequently input into a Model Predictive Control (MPC) model as reference trajectories. MPC predicts the vehicle’s behavior over a future time horizon and computes the optimal control inputs (such as acceleration and steering angle), which are fed back into the vehicle control system to dynamically generate realistic driving behavior. The MPC control model not only enhances the vehicle's path-tracking performance but also ensures the physical feasibility of the dynamic behavior, thereby enabling the autonomous driving system to exhibit greater adaptability and robustness in complex scenarios \cite{stano2023model}. MPC is an online optimal control strategy that achieves optimal decision-making in dynamic environments through prediction of the system’s future behavior.\par

The optimization objective of our MPC is to guide the vehicle state to closely follow the reference trajectory by utilizing the input set \(way_k\), while minimizing the cost associated with control inputs. The objective function is defined as follows:
\begin{equation}
J = \sum_{k=0}^{N} \left[ \left( s_k^{mpc} - way_k \right)^T V \left( s_k^{mpc} - way_k \right) + u_k^T W u_k \right]
\end{equation}
where, \( s_k^{\text{mpc}} \) denotes the system state vector, which includes the vehicle's position, velocity, acceleration, and other relevant states. \( u_k \) represents the control input vector, and \( \text{waypoint}_k \) denotes the reference trajectory points. \( V \) and \( W \) are weighting matrices that quantify the contributions of the state deviation and control effort to the objective function, respectively. \( N \) is the length of the prediction horizon, which is set to 5 by default.\par
By incorporating the objective function with the constraint conditions, MPC formulates the problem as a constrained quadratic programming (QP) problem, which is then solved using CVXPY.
\begin{equation}
s_{\text{min}}^{\text{mpc}} \leq s_k^{\text{mpc}} \leq s_{\text{max}}^{\text{mpc}}
\end{equation}
\begin{equation}
u_{\text{min}} \leq u_k \leq u_{\text{max}}
\end{equation}
\begin{equation}
\Delta u_{\text{min}} \leq u_k - u_{k-1} \leq \Delta u_{\text{max}}
\end{equation}
\begin{equation}
\min_{u_0, \ldots, u_{N-1}} J \quad \text{s.t. constraints}
\end{equation} \par
After solving the optimization problem, MPC extracts only the current control input \( u_t = u_0^* \) and applies it to the vehicle control system, advancing the system to the next time step. Subsequently, the control output from MPC is executed within the Carla simulation environment, where the vehicle responds to the control commands and the simulation results are fed back. The environment then computes the reward based on the current scenario, which serves as a crucial signal for the optimizer to refine the VAA model. This entire process forms a closed-loop system that iteratively improves performance and stability in complex driving scenarios through continuous decision-making, control, simulation, and optimization.\par

\section{Experiment}

The experimental setup utilizes a Windows 11 64-bit operating system, equipped with an NVIDIA GeForce RTX 4090 GPU, an Intel(R) Core(TM) i7-12700KF CPU @ 2.10GHz, and 32.0GB of RAM. All code is implemented in Python 3.8. The scenario generation and simulation platform is built upon Carla version 0.9.13. In this study, the key features of emergency lane-changing behavior include the vehicle's speed and lateral position. The simulation scenarios are initialized in Carla, where the attacking vehicle adopts an emergency lane-change strategy based on its relative position to the target vehicle. The initialized scenario terminates upon meeting specific conditions, such as reaching a simulation duration of 10 seconds or encountering a collision between vehicles. \par

\subsection{Emergency Lane Change Trajectory Learning}
In the generator, the greater the similarity between the generated lane-changing trajectories and those obtained from real-world data, the more accurate the generation algorithm is considered. A commonly used evaluation method, Maximum Likelihood Estimation (MLE), minimizes the cross-entropy between the real data distribution and the generated data, expressed as \( -E_{y \sim p} \log q(y) \). However, the most accurate way to evaluate a generative model is to sample from it and have human observers assess the generated data based on prior knowledge \cite{yu2017seqgan}. Accordingly, we use a dataset of 520 lane-change samples as the real data distribution \( p(y_i \mid y_1, \ldots, y_{520}) \). The real data distribution \( G_{\text{real}} \) can thus be regarded as representative of human observation in the real world, and a more appropriate evaluation metric should be:
\begin{equation}
NLL_{\text{real}} = -E_{Y_{1:T} \sim G_\theta} \left[ \sum_{t=1}^{T} \log G_{\text{real}}(y_t \mid Y_{1:t-1}) \right]
\end{equation} \par
During the testing phase, we use \( G_\theta \) to generate 50,000 sequence samples and compute the Negative Log-Likelihood with respect to the real data distribution, \( NLL_{\text{real}} \), for each sample based on \( G_{\text{real}} \) and its average score. We further compare the results with baseline methods and SeqGAN. To evaluate the contribution of each optimized component, we also conduct ablation studies on our proposed modules.\par
\begin{table}[htbp]
\centering
\caption{Dissolution experiment results}
\label{tab:1}
\resizebox{0.48\textwidth}{!}{%
\begin{tabular}{lccc}
\toprule
\textbf{$Loss$} & \textbf{$Seqgan$} & \textbf{$Seqgan\_GRU$} & \textbf{$Seqgan\_GRU\_Attention$} \\
\midrule
NLL & 11.235 & 10.371 & 10.112 \\
\bottomrule
\end{tabular}%
}
\end{table}
\begin{figure}
	\centering 
	\includegraphics[width=0.42\textwidth]{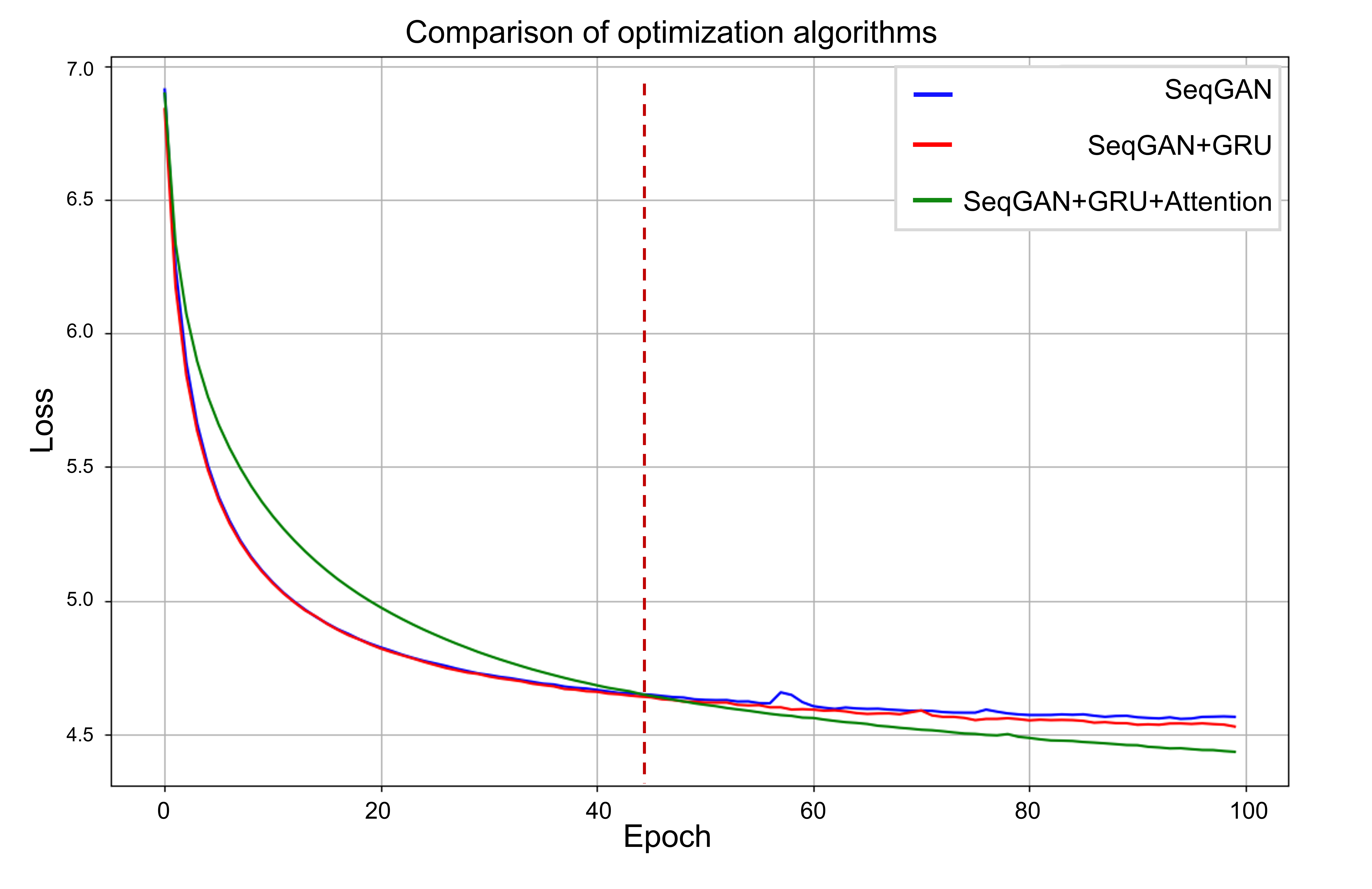}	
	\caption{The Triggering Process of the Emergency Lane Change Strategy  } 
	\label{6}%
\end{figure}
As illustrated in Figure \ref{6} and Table \ref{tab:1}, the integration of the proposed modules significantly enhances the trajectory learning capabilities of the adversarial network. Each added component contributes positively, collectively leading to improved performance.
In addition, we randomly selecte 50 sample data points from the dataset for training purposes and systematically compared the sequence generation performance of three methods: planned sampling, PG-BLEU, and SeqGAN.\par
\begin{table}[htbp]
\centering
\caption{Comparative results of sequence generation performance}
\label{tab:2}
\resizebox{0.48\textwidth}{!}{%
\begin{tabular}{lcccc}
\toprule
\textbf{$Loss$} & \textbf{$SS$} & \textbf{$PG-BLEU$} & \textbf{$Seqgan$} & \textbf{$GA\_Seqgan$} \\
\midrule
NLL & 14.431 & 14.242 & 13.143 & 11.451 \\
p-value & $<10^{-6}$ & $<10^{-6}$ & -- & -- \\
\bottomrule
\end{tabular}%
}
\end{table}
Furthermore, as illustrated in the results presented in Table \ref{tab:2}, our GA-SeqGAN demonstrates superior performance in sequence generation tasks with a limited number of samples while maintaining minimal deviation from the actual data distribution. These findings suggest that the trained generator has effectively learned lane-changing behaviors closely resembling human behaviors, thus establishing a robust foundation for subsequent behavioral guidance applications.\par

\subsection{Assessment of Strategies for Active Attack Behaviors}
We evaluate the proposed proactive vehicle attack behavior strategy by comparing control policies with and without the integration of BLM. In the parameter settings of the attack strategy, we adopt a learning rate of 0.0003 to ensure the stability of model convergence, and accumulate 1280 time steps to serve as the basis for policy updates. Each training batch consists of 128 samples, balancing training efficiency and memory usage. The model is updated for 20 iterations per training cycle to enhance policy fitting capability. An entropy regularization coefficient of 0.01 is used to encourage policy diversity, and a discount factor of 0.99 is set to ensure long-term reward consideration, enabling fine-grained optimization of the attack strategy.The weights for \( \lambda_1 \) to \( \lambda_5 \) are set as 1, 0.25, 0.25, 5, and 50, respectively.\par

\begin{figure}
	\centering 
	\includegraphics[width=0.48\textwidth]{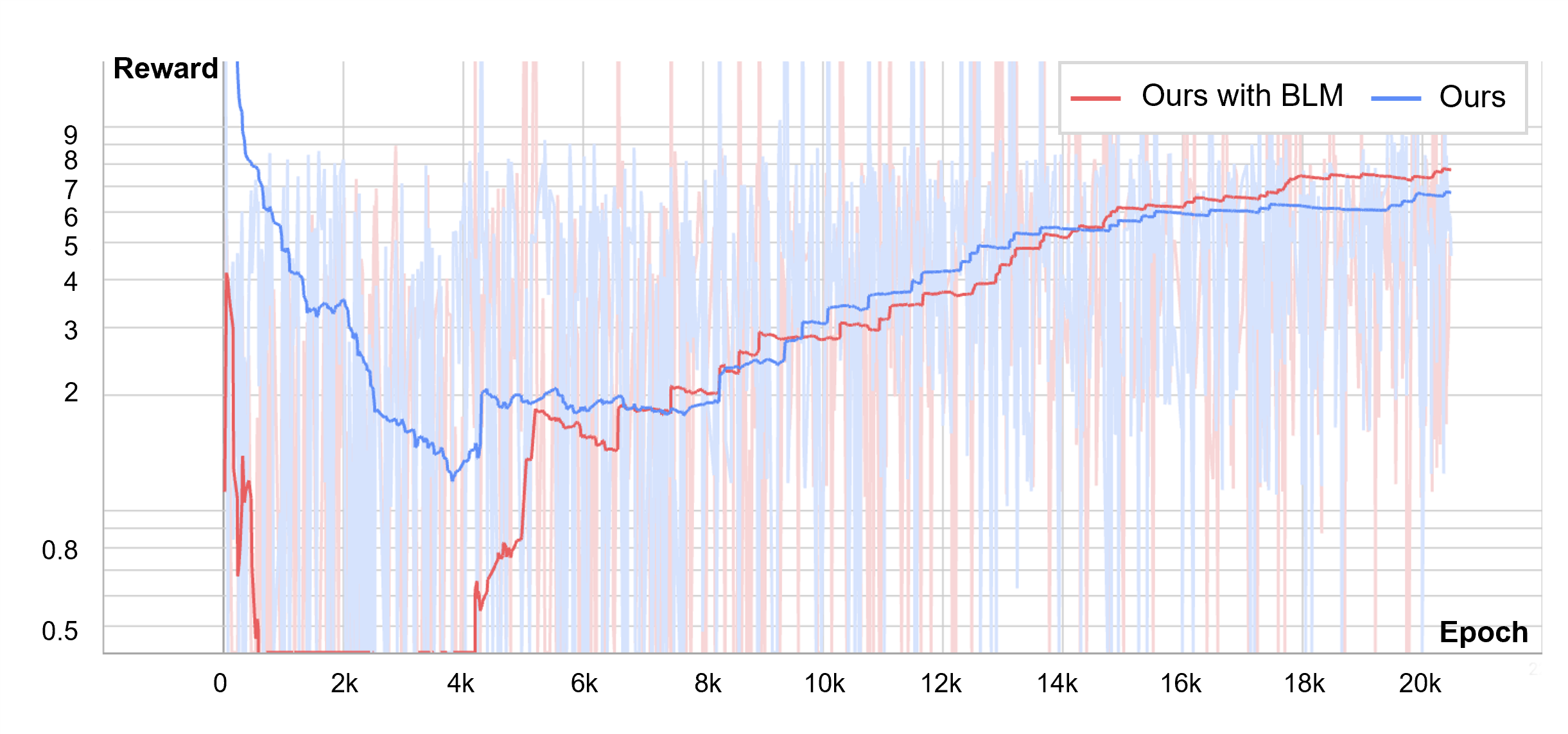}	
	\caption{Training Reward Curve (Dark Line: Mean Reward)} 
	\label{7}%
\end{figure}

\begin{figure}
	\centering 
	\includegraphics[width=0.48\textwidth]{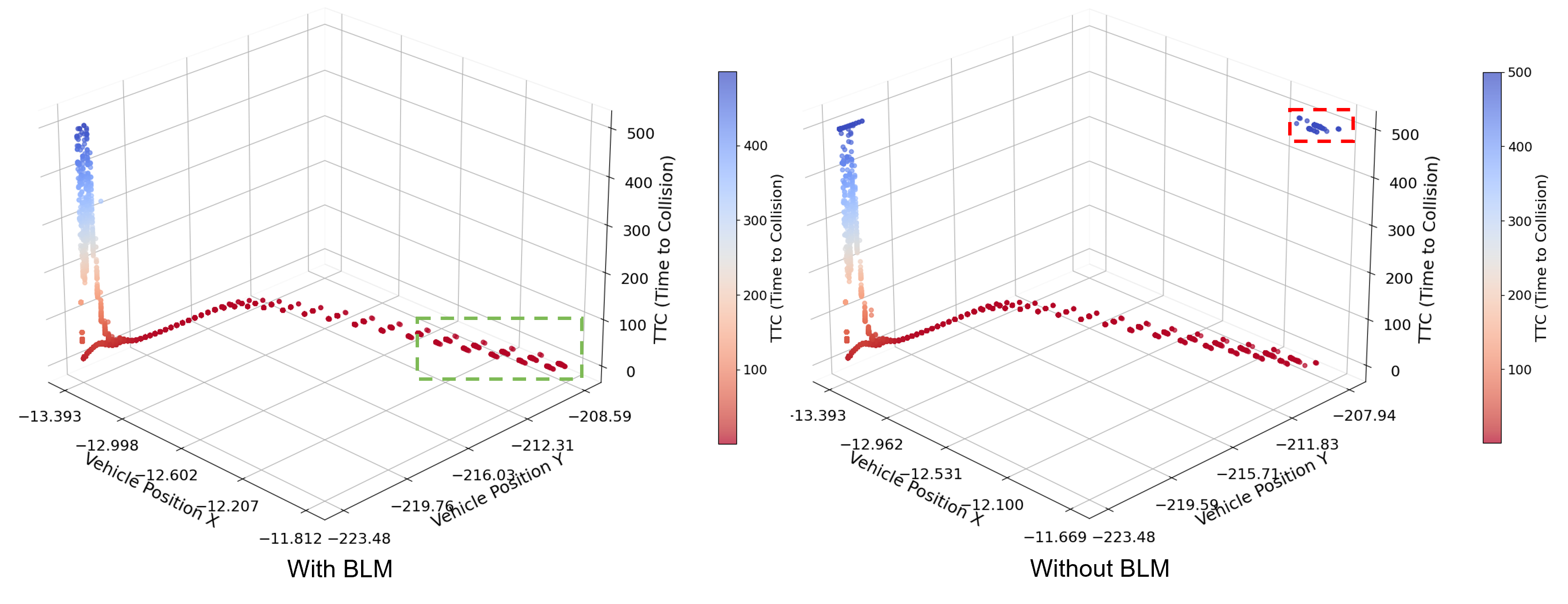}	
	\caption{TTC Variation with Spatial Exploration} 
	\label{8}%
\end{figure}

As shown in figures 7 and 8, introducing the BLM module significantly improves the efficiency and stability of reward convergence during training. In Figure \ref{7}, the reward curve reveals that the model, with BLM, maintains its capability to explore high-risk areas while exhibiting reduced fluctuation, a smoother trend, and ultimately reaches a higher and more stable reward value. Furthermore, Figure \ref{8} illustrates the spatial distribution of Time-To-Collision (TTC) over 100 simulation trials, comparing the results with and without BLM. In scenarios without BLM, regions with high TTC values are more densely distributed, indicating a lower probability of collisions. It is also observed that vehicles without BLM tend to generate dense clusters of blue points after lane changes, representing relatively safer positions relative to surrounding vehicles.These results indicate that the inclusion of BLM not only enhances reward performance and stabilizes risk-exploration behavior for the ego vehicle, but also improves the overall safety and robustness of the vehicle’s behavior strategy.\par
To validate the rationality of our design of various influencing factors within the reward function \(R_t\), we adopt SHAP analysis to evaluate the contribution of different indicators to the model's convergence \cite{van2022tractability}. SHAP values are derived from the Shapley values in cooperative game theory, which fairly allocate the total gain among participants based on their individual contributions to the overall outcome. The SHAP value \(\phi\) is analyzed using the following formulation:
\begin{equation}
\phi_j = \sum_{S \subseteq F \setminus \{j\}} \frac{|S|! \, (|F| - |S| - 1)!}{|F|!} \left[ f_x(S \cup \{j\}) - f_x(S) \right]\
\end{equation}
where, \(F\) denotes the set of all features, \(S\) represents a subset that does not contain feature \(j\), and \(f_x(S)\) is the model output when only features in subset \(S\) are present. \(|S|\) is the number of features in subset \(S\), and \(|F|\) is the total number of features. \par
\begin{figure}
	\centering 
	\includegraphics[width=0.48\textwidth]{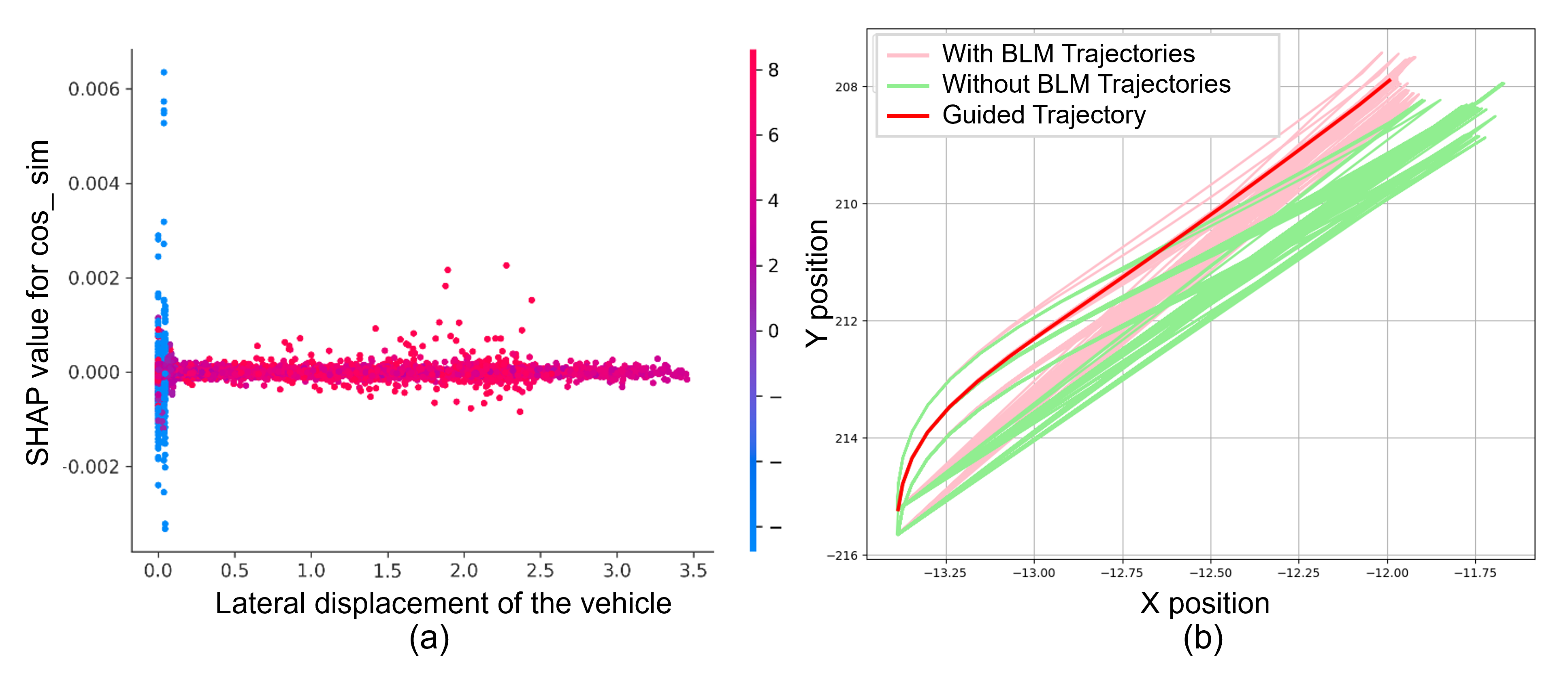}	
	\caption{(a) The Contribution of cos-sim to Vehicle Attack Strategy;(b) Emergency lane-change behavior of the attacking vehicle with and without behavior guidance} 
	\label{9}%
\end{figure}

The SHAP value represents the average marginal contribution of feature \(j\) across all possible feature combinations. Experimental results demonstrate that our BLM, when combined with the cosine function, can effectively mitigate vehicle collision attacks, as shown in the Figure \ref{9} (a). Figure \ref{9} (b) presents a comparison of the trajectory distributions of ego vehicles with and without the Behavioral Learning Model (BLM) across 100 experimental trials. In this figure, the red line denotes the behavioral guidance trajectory generated by BLM, while the pink and green lines correspond to the trajectory distributions in the absence and presence of behavioral guidance, respectively. It is evident that upon enabling behavioral guidance, the generated attack trajectories align more closely with those derived from real-world data within the BLM framework. This finding demonstrates that the proposed strategy successfully integrates BLM and effectively guides the emergency lane-changing behavior of the model.\par
\begin{figure}
	\centering 
	\includegraphics[width=0.48\textwidth]{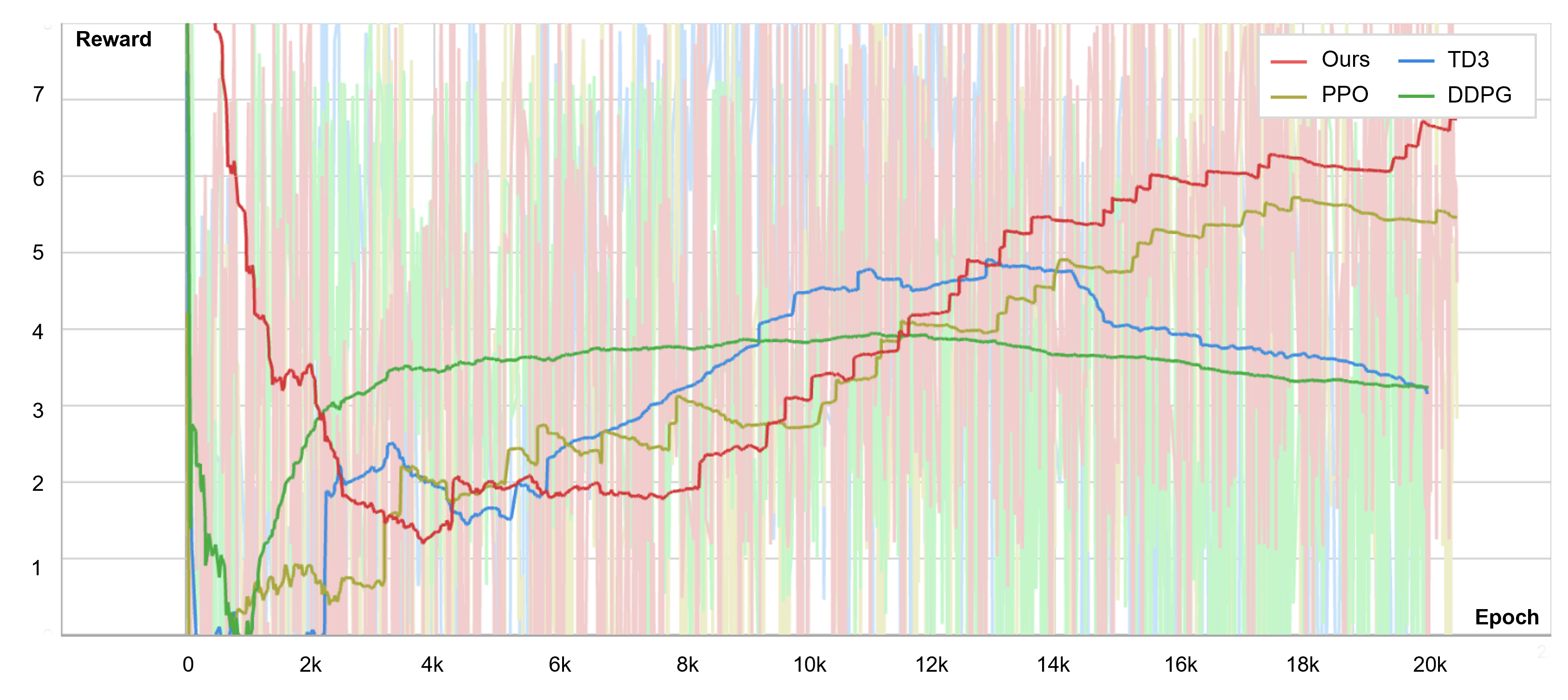}	
	\caption{Analysis of Model Training Convergence Under Different Strategies} 
	\label{10}%
\end{figure}

Similarly, we evaluate the superiority of our control method by comparing it with different baseline control strategies. As shown in the Figure \ref{10}, our approach our demonstrates significantly faster convergence and higher stability in reward values compared to PPO, TD3, and DDPG after a certain number of training iterations. Notably, our method exhibits higher learning efficiency in the early training phase and achieves superior reward performance during the later convergence stage. These results further indicate that our control strategy effectively optimizes decision-making and enhances overall system performance.\par
\begin{figure}
	\centering 
	\includegraphics[width=0.47\textwidth]{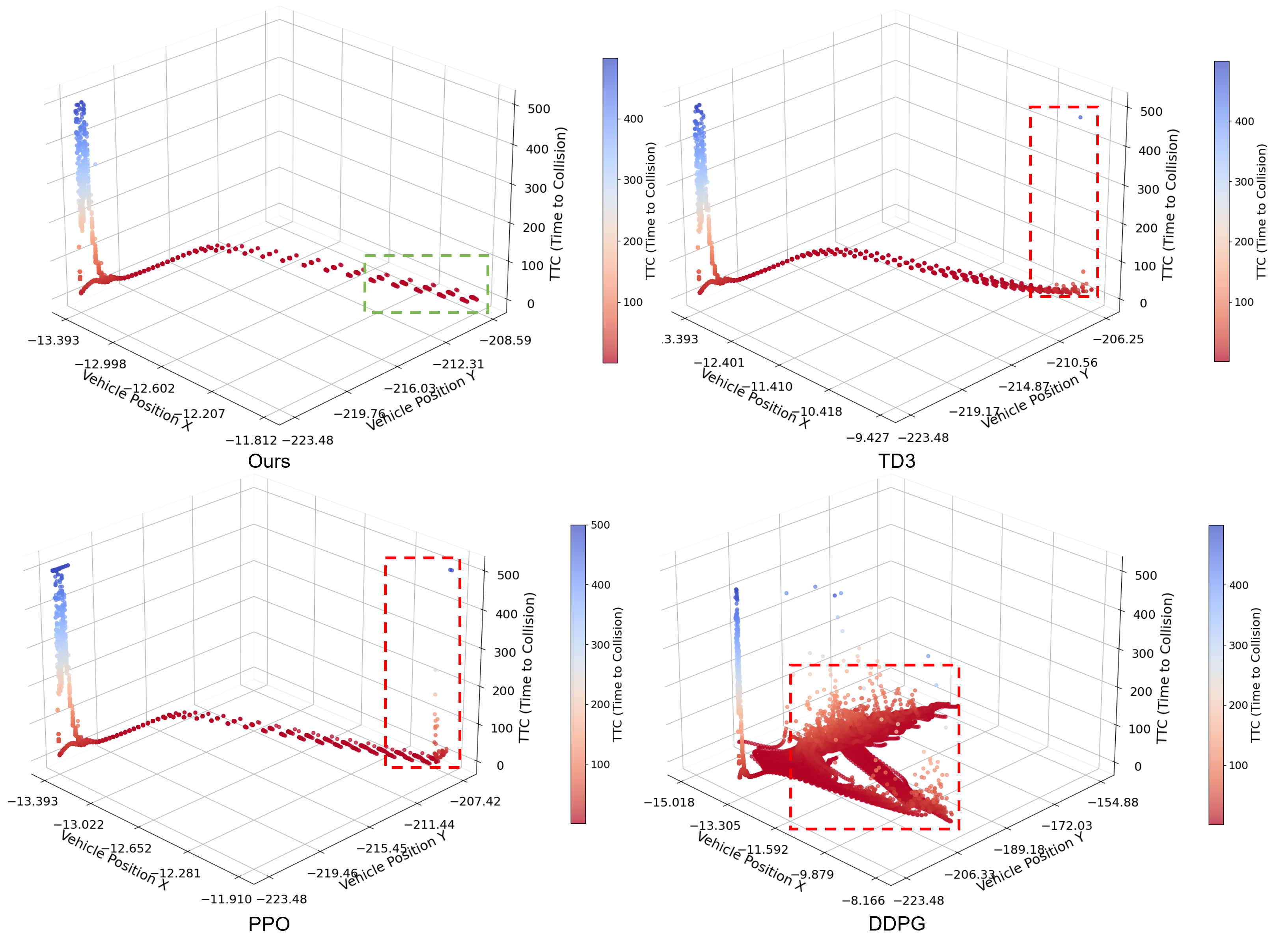}	
	\caption{TTC variation with spatial exploration under different strategies} 
	\label{11}%
\end{figure}

The results show that our method demonstrates a clear advantage in proactive attack strategies. Figure \ref{11} illustrates the 3D distribution of TTC values under different strategies as spatial exploration varies. It can be observed that our strategy (Ours) enables the model to more precisely control vehicle behaviors and explore critical hazardous regions. In contrast, traditional methods (e.g., TD3, PPO, and DDPG) tend to exhibit uneven or overly dispersed exploration patterns. This indicates that our method not only better achieves attack objectives but also adapts to complex scenarios in a more stable and efficient manner, further validating its potential in autonomous driving safety testing and optimization.\par

\subsection{Analysis of Emergency Lane-Change Collision Scenario Generation}

\begin{figure}
	\centering 
	\includegraphics[width=0.44\textwidth]{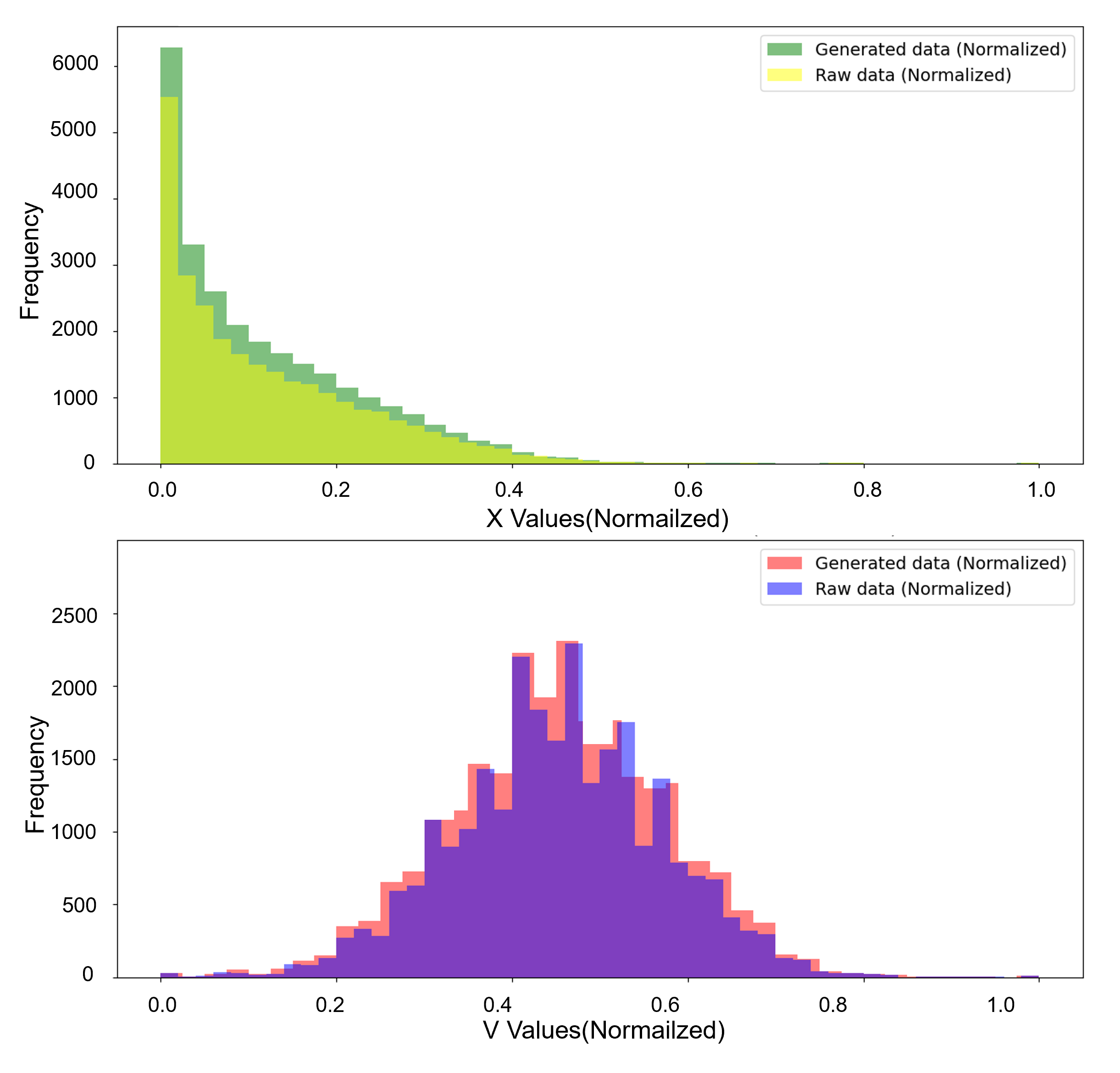}	
	\caption{Comparison of generated data and original data X, V distributions (X and V denote the lateral displacement and longitudinal velocity of the vehicle, respectively.)} 
	\label{12}%
\end{figure}

In the Carla simulation environment, we align the simulation frame rate with that of the HighD dataset. To ensure that the generated trajectories successfully capture emergency lane-change behaviors, it is essential to verify whether the generated scenarios exhibit vehicle behavior consistent with the dataset. For this purpose, we compare the generated scenario trajectories with those from the dataset in terms of lateral displacement and longitudinal velocity. As shown in Figure \ref{12}, the distribution of generated data aligns closely with the distribution of real data, especially in low-value regions, where both exhibit similar frequency patterns. This suggests that our generative model accurately captures the key characteristics of real-world lane-change behaviors, thereby providing a reliable foundation for downstream applications.\par
\begin{figure}
	\centering 
	\includegraphics[width=0.46\textwidth]{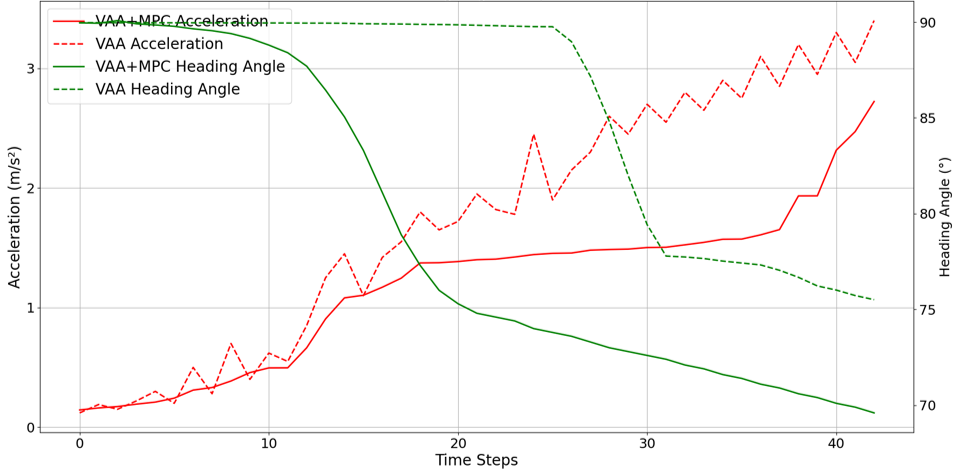}	
	\caption{Analysis of the Impact of adding MPC on the VAA Model} 
	\label{13}%
\end{figure}

As shown in Figure~\ref{13}, the VAA+MPC approach (solid lines) yields more physically realistic vehicle dynamics compared to VAA alone (dashed lines). The acceleration profile (red) demonstrates that VAA+MPC maintains smoother transitions within feasible limits (0--3 m/s\textsuperscript{2}), whereas VAA exhibits noticeable erratic fluctuations. Similarly, the heading angle (green) indicates that VAA+MPC enables a controlled and continuous turning maneuver after a period of stable orientation, in contrast to VAA, which presents premature and inconsistent steering adjustments. These results highlight the effectiveness of MPC in enforcing physical constraints and applying predictive smoothing to achieve stable and reliable autonomous vehicle control.

\begin{figure}
	\centering 
	\includegraphics[width=0.46\textwidth]{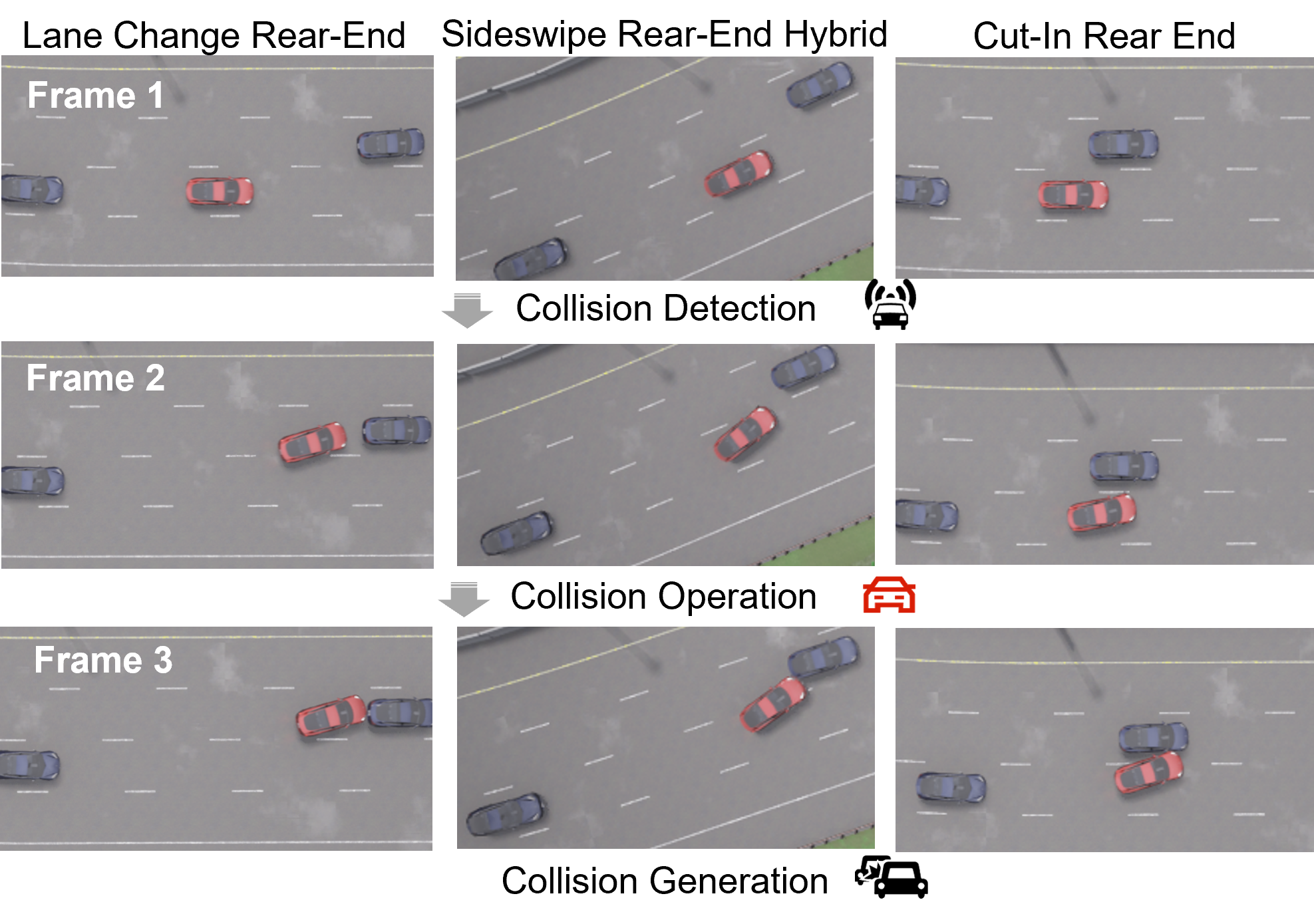}	
	\caption{Visualization of .emergency lane-change collision} 
	\label{14}%
\end{figure}

Leveraging the Carla simulation platform, our scenario generation framework is capable of capturing multi-view data representations, including bird’s-eye view (BEV) perspectives. The generated risky scenarios are illustrated in Figure \ref{14}. In addition, sensor data can be utilized to capture the driver’s perspective from the impacted vehicle, as well as point cloud data, thereby facilitating the construction of a comprehensive hazardous driving dataset.\par
\begin{figure}
	\centering 
	\includegraphics[width=0.48\textwidth]{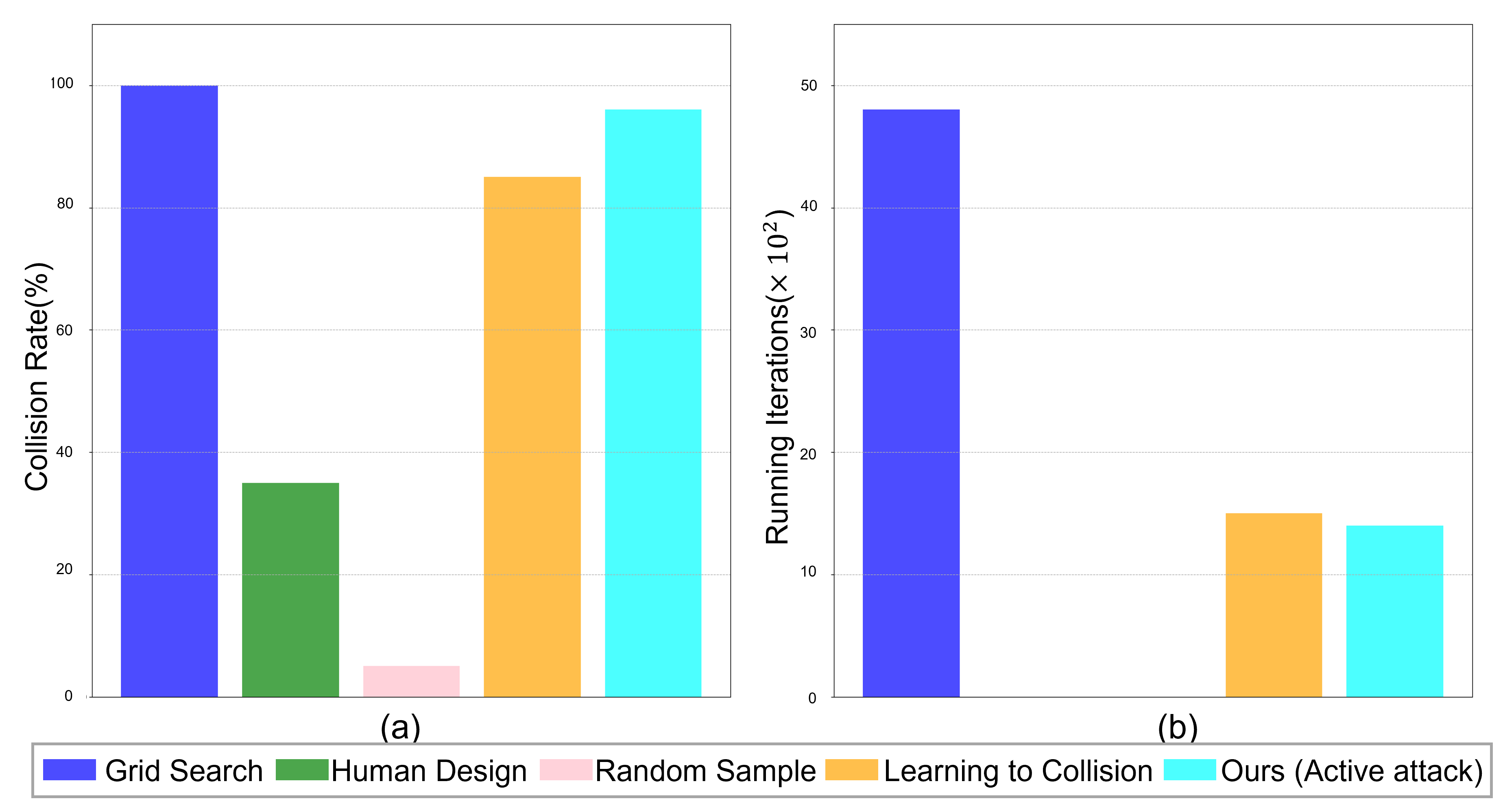}	
	\caption{(a) Comparison of collision success rates achieved by different methods; (b) Comparison of computational time required by different methods} 
	\label{15}%
\end{figure}

Finally, we establish four baselines for comparison: 1) Grid search: This represents the most straightforward approach to searching for risk. Following the methodology in \cite{zhao2016accelerated}, we discretize the vehicle strategy and exhaustively search all possible combinations to identify the optimal solution. 2) Manual design: We adopt the same parameters and programs used in the Carla Challenge competition \cite{cui2019class} as our baseline. 3) Random sampling: Strategies are sampled uniformly from a uniform distribution. 4) Learning to collision: Proposed by Ding et al. \cite{ding2020learning}, this method employs a learning-based approach. We evaluate its performance using the collision success rate after the model reached a stable strategy, as well as the number of iterations required for stabilization. The results are presented in Figure \ref{15}. The findings indicate that the active attack strategy significantly outperforms other benchmark methods in terms of collision success rate while maintaining a relatively low computational cost. \par
\section{Conclusion}

This paper proposes a method for generating risky and emergency lane-changing scenarios based on behavior learning. First, a Behavior Learning Module (BLM) based on an improved GA-SeqGAN is employed to learn from a limited dataset of emergency lane-changing behaviors. This approach enables the generation of more diverse and realistic vehicle maneuvers. Subsequently, the ego vehicle is modeled as an agent, while the surrounding environment, including roads and other vehicles, is treated as the external environment. When a dangerous scenario is successfully generated, the agent receives a high reward. The entire process is optimized using an iterative strategy that approximates a deep reinforcement learning framework, and is integrated with the behavior learning module to simulate human-like emergency lane-changing attacks.\par
Experimental results demonstrate the effectiveness of the proposed method in generating emergency lane-change collision scenarios, with the following key advantages:\par
1.The emergency lane-changing model effectively learns emergency trajectories from a small dataset while ensuring high data diversity.\par
2.The integration of Recurrent Proximal Policy Optimization (RPPO) and Model Predictive Control (MPC) ensures that vehicle behaviors comply with physical constraints, while enhancing adaptability to environmental parameters and exploration of risk-inducing situations.\par
3.Compared to benchmark methods, the proposed active attack strategy increases the likelihood of collision and scenario generation, all while maintaining a low computational cost.\par
In future work, vehicles equipped with the proposed attack strategy will be tested in interactions with autonomous vehicles employing state-of-the-art driving algorithms. This will allow for the evaluation of vulnerabilities in current autonomous driving systems and support the development of improved safety mechanisms. Additionally, we plan to convert the generated trajectory data into realistic video sequences and radar point cloud data for target detection and accident analysis. These data can not only validate the realism of the generated scenarios, but also serve as valuable input for enhancing autonomous driving systems and advancing research in traffic safety.

%% The Appendices part is started with the command \appendix;
%% appendix sections are then done as normal sections

% Can use something like this to put references on a page
% by themselves when using endfloat and the captionsoff option.
\ifCLASSOPTIONcaptionsoff
  \newpage
\fi

% trigger a \newpage just before the given reference
% number - used to balance the columns on the last page
% adjust value as needed - may need to be readjusted if
% the document is modified later
%\IEEEtriggeratref{8}
% The "triggered" command can be changed if desired:
%\IEEEtriggercmd{\enlargethispage{-5in}}

% references section

% can use a bibliography generated by BibTeX as a .bbl file
% BibTeX documentation can be easily obtained at:
% http://mirror.ctan.org/biblio/bibtex/contrib/doc/
% The IEEEtran BibTeX style support page is at:
% http://www.michaelshell.org/tex/ieeetran/bibtex/
%\bibliographystyle{IEEEtran}
% argument is your BibTeX string definitions and bibliography database(s)
%\bibliography{IEEEabrv,../bib/paper}
%
% <OR> manually copy in the resultant .bbl file
% set second argument of \begin to the number of references
% (used to reserve space for the reference number labels box)
\bibliographystyle{IEEEtran}
\bibliography{IEEEabrv, re}

@article{kalra2016driving,
  title={Driving to safety: How many miles of driving would it take to demonstrate autonomous vehicle reliability?},
  author={Kalra, Nidhi and Paddock, Susan M},
  journal={Transportation Research Part A: Policy and Practice},
  volume={94},
  pages={182--193},
  year={2016},
  publisher={Elsevier}
}

@inproceedings{cui2019class,
  title={Class-balanced loss based on effective number of samples},
  author={Cui, Yin and Jia, Menglin and Lin, Tsung-Yi and Song, Yang and Belongie, Serge},
  booktitle={Proceedings of the IEEE/CVF conference on computer vision and pattern recognition},
  pages={9268--9277},
  year={2019}
}

@article{abdel2024matched,
  title={A matched case-control analysis of autonomous vs human-driven vehicle accidents},
  author={Abdel-Aty, Mohamed and Ding, Shengxuan},
  journal={Nature communications},
  volume={15},
  number={1},
  pages={4931},
  year={2024},
  publisher={Nature Publishing Group UK London}
}

@article{zhao2016accelerated,
  title={Accelerated evaluation of automated vehicles safety in lane-change scenarios based on importance sampling techniques},
  author={Zhao, Ding and Lam, Henry and Peng, Huei and Bao, Shan and LeBlanc, David J and Nobukawa, Kazutoshi and Pan, Christopher S},
  journal={IEEE transactions on intelligent transportation systems},
  volume={18},
  number={3},
  pages={595--607},
  year={2016},
  publisher={IEEE}
}

@misc{nhtsa2018fars,
  title        = {Fatality Analysis Reporting System (FARS) 2018 Data},
  author       = {{National Highway Traffic Safety Administration}},
  year         = {2019},
  howpublished = {\url{https://www.nhtsa.gov/research-data/fatality-analysis-reporting-system-fars}},
  note         = {Accessed: 2025-04-26}
}

@article{rossi2021vehicle,
  title={Vehicle trajectory prediction and generation using LSTM models and GANs},
  author={Rossi, Luca and Ajmar, Andrea and Paolanti, Marina and Pierdicca, Roberto},
  journal={Plos one},
  volume={16},
  number={7},
  pages={e0253868},
  year={2021},
  publisher={Public Library of Science San Francisco, CA USA}
}

@inproceedings{xie2024advdiffuser,
  title={AdvDiffuser: Generating Adversarial Safety-Critical Driving Scenarios via Guided Diffusion},
  author={Xie, Yuting and Guo, Xianda and Wang, Cong and Liu, Kunhua and Chen, Long},
  booktitle={2024 IEEE/RSJ International Conference on Intelligent Robots and Systems (IROS)},
  pages={9983--9989},
  year={2024},
  organization={IEEE}
}

@inproceedings{fremont2020formal,
  title={Formal scenario-based testing of autonomous vehicles: From simulation to the real world},
  author={Fremont, Daniel J and Kim, Edward and Pant, Yash Vardhan and Seshia, Sanjit A and Acharya, Atul and Bruso, Xantha and Wells, Paul and Lemke, Steve and Lu, Qiang and Mehta, Shalin},
  booktitle={2020 IEEE 23rd International Conference on Intelligent Transportation Systems (ITSC)},
  pages={1--8},
  year={2020},
  organization={IEEE}
}

@article{zhong2021survey,
  title={A survey on scenario-based testing for automated driving systems in high-fidelity simulation},
  author={Zhong, Ziyuan and Tang, Yun and Zhou, Yuan and Neves, Vania de Oliveira and Liu, Yang and Ray, Baishakhi},
  journal={arXiv preprint arXiv:2112.00964},
  year={2021}
}

@article{cai2022survey,
  title={A survey on data-driven scenario generation for automated vehicle testing},
  author={Cai, Jinkang and Deng, Weiwen and Guang, Haoran and Wang, Ying and Li, Jiangkun and Ding, Juan},
  journal={Machines},
  volume={10},
  number={11},
  pages={1101},
  year={2022},
  publisher={MDPI}
}

@article{tang2023survey,
  title={A survey on automated driving system testing: Landscapes and trends},
  author={Tang, Shuncheng and Zhang, Zhenya and Zhang, Yi and Zhou, Jixiang and Guo, Yan and Liu, Shuang and Guo, Shengjian and Li, Yan-Fu and Ma, Lei and Xue, Yinxing and others},
  journal={ACM Transactions on Software Engineering and Methodology},
  volume={32},
  number={5},
  pages={1--62},
  year={2023},
  publisher={ACM New York, NY}
}

@inproceedings{zhang2023novel,
  title={A novel scenario-based testing approach for cooperative-automated driving systems},
  author={Zhang, Xizhe and Mo, Yuen Kwan and Chodowiec, Emil and Tang, Yun and Higgins, Matthew and Khastgir, Siddartha and Jennings, Paul},
  booktitle={2023 IEEE International Conference on Systems, Man, and Cybernetics (SMC)},
  pages={3487--3494},
  year={2023},
  organization={IEEE}
}

@article{zhou2024would,
  title={How would autonomous vehicles behave in real-world crash scenarios?},
  author={Zhou, Rui and Zhang, Guoqing and Huang, Helai and Wei, Zhiyuan and Zhou, Hanchu and Jin, Jielin and Chang, Fangrong and Chen, Jiguang},
  journal={Accident Analysis \& Prevention},
  volume={202},
  pages={107572},
  year={2024},
  publisher={Elsevier}
}

@article{schwall2020waymo,
  title={Waymo public road safety performance data},
  author={Schwall, Matthew and Daniel, Tom and Victor, Trent and Favaro, Francesca and Hohnhold, Henning},
  journal={arXiv preprint arXiv:2011.00038},
  year={2020}
}

@article{feng2021intelligent,
  title={Intelligent driving intelligence test for autonomous vehicles with naturalistic and adversarial environment},
  author={Feng, Shuo and Yan, Xintao and Sun, Haowei and Feng, Yiheng and Liu, Henry X},
  journal={Nature communications},
  volume={12},
  number={1},
  pages={748},
  year={2021},
  publisher={Nature Publishing Group UK London}
}

@article{hao2023bridging,
  title={Bridging Data-Driven and Knowledge-Driven Approaches for Safety-Critical Scenario Generation in Automated Vehicle Validation},
  author={Hao, Kunkun and Liu, Lu and Cui, Wen and Zhang, Jianxing and Yan, Songyang and Pan, Yuxi and Yang, Zijiang},
  journal={arXiv preprint arXiv:2311.10937},
  year={2023}
}

@inproceedings{xu2023diffscene,
  title={Diffscene: Diffusion-based safety-critical scenario generation for autonomous vehicles},
  author={Xu, Chejian and Zhao, Ding and Sangiovanni-Vincentelli, Alberto and Li, Bo},
  booktitle={The Second Workshop on New Frontiers in Adversarial Machine Learning},
  year={2023}
}

@article{feng2023dense,
  title={Dense reinforcement learning for safety validation of autonomous vehicles},
  author={Feng, Shuo and Sun, Haowei and Yan, Xintao and Zhu, Haojie and Zou, Zhengxia and Shen, Shengyin and Liu, Henry X},
  journal={Nature},
  volume={615},
  number={7953},
  pages={620--627},
  year={2023},
  publisher={Nature Publishing Group UK London}
}

@inproceedings{yu2017seqgan,
  title={Seqgan: Sequence generative adversarial nets with policy gradient},
  author={Yu, Lantao and Zhang, Weinan and Wang, Jun and Yu, Yong},
  booktitle={Proceedings of the AAAI conference on artificial intelligence},
  volume={31},
  number={1},
  year={2017}
}

@inproceedings{dosovitskiy2017carla,
  title={CARLA: An open urban driving simulator},
  author={Dosovitskiy, Alexey and Ros, German and Codevilla, Felipe and Lopez, Antonio and Koltun, Vladlen},
  booktitle={Conference on robot learning},
  pages={1--16},
  year={2017},
  organization={PMLR}
}

@article{rafiei2022pedestrian,
  title={Pedestrian collision avoidance using deep reinforcement learning},
  author={Rafiei, Alireza and Fasakhodi, Amirhossein Oliaei and Hajati, Farshid},
  journal={International journal of automotive technology},
  volume={23},
  number={3},
  pages={613--622},
  year={2022},
  publisher={Springer}
}

@inproceedings{tan2021scenegen,
  title={Scenegen: Learning to generate realistic traffic scenes},
  author={Tan, Shuhan and Wong, Kelvin and Wang, Shenlong and Manivasagam, Sivabalan and Ren, Mengye and Urtasun, Raquel},
  booktitle={Proceedings of the IEEE/CVF Conference on Computer Vision and Pattern Recognition},
  pages={892--901},
  year={2021}
}

@inproceedings{jenkins2018accident,
  title={Accident scenario generation with recurrent neural networks},
  author={Jenkins, Ian Rhys and Gee, Ludvig Oliver and Knauss, Alessia and Yin, Hang and Schroeder, Jan},
  booktitle={2018 21st International Conference on Intelligent Transportation Systems (ITSC)},
  pages={3340--3345},
  year={2018},
  organization={IEEE}
}

@article{li2024autonomous,
  title={Autonomous Driving Decision Algorithm for Complex Multi-Vehicle Interactions: An Efficient Approach Based on Global Sorting and Local Gaming},
  author={Li, Daofei and Zhang, Jiajie and Liu, Guanming},
  journal={IEEE Transactions on Intelligent Transportation Systems},
  year={2024},
  publisher={IEEE}
}

@article{hoel2019combining,
  title={Combining planning and deep reinforcement learning in tactical decision making for autonomous driving},
  author={Hoel, Carl-Johan and Driggs-Campbell, Katherine and Wolff, Krister and Laine, Leo and Kochenderfer, Mykel J},
  journal={IEEE transactions on intelligent vehicles},
  volume={5},
  number={2},
  pages={294--305},
  year={2019},
  publisher={IEEE}
}

@article{lu2022learning,
  title={Learning configurations of operating environment of autonomous vehicles to maximize their collisions},
  author={Lu, Chengjie and Shi, Yize and Zhang, Huihui and Zhang, Man and Wang, Tiexin and Yue, Tao and Ali, Shaukat},
  journal={IEEE Transactions on Software Engineering},
  volume={49},
  number={1},
  pages={384--402},
  year={2022},
  publisher={IEEE}
}

@article{qin2019automatic,
  title={Automatic testing with reusable adversarial agents},
  author={Qin, Xin and Ar{\'e}chiga, Nikos and Best, Andrew and Deshmukh, Jyotirmoy},
  journal={arXiv preprint arXiv:1910.13645},
  year={2019}
}

@article{demetriou2023deep,
  title={A deep learning framework for generation and analysis of driving scenario trajectories},
  author={Demetriou, Andreas and Alfsv{\aa}g, Henrik and Rahrovani, Sadegh and Haghir Chehreghani, Morteza},
  journal={SN Computer Science},
  volume={4},
  number={3},
  pages={251},
  year={2023},
  publisher={Springer}
}

@article{jia2023conditional,
  title={Conditional temporal GAN for intent-aware vessel trajectory prediction in the precautionary area},
  author={Jia, Chengfeng and Ma, Jie},
  journal={Engineering Applications of Artificial Intelligence},
  volume={126},
  pages={106776},
  year={2023},
  publisher={Elsevier}
}

@inproceedings{han2021variable,
  title={Variable speed trajectory prediction method of dynamic prosthesis based on improved SeqGAN},
  author={Han, Haowei and An, Honglei and Ma, Hongxu and Wei, Qing},
  booktitle={2021 China Automation Congress (CAC)},
  pages={2628--2632},
  year={2021},
  organization={IEEE}
}

@article{acquarone2023cooperative,
  title={Cooperative Adaptive Cruise Control Based on Reinforcement Learning for Heavy-Duty BEVs},
  author={Acquarone, Matteo and Miretti, Federico and Misul, Daniela and Sassara, Luca},
  journal={IEEE Access},
  volume={11},
  pages={127145--127156},
  year={2023},
  publisher={IEEE}
}

@article{siboo2023empirical,
  title={An empirical study of ddpg and ppo-based reinforcement learning algorithms for autonomous driving},
  author={Siboo, Sanjna and Bhattacharyya, Anushka and Raj, Rashmi Naveen and Ashwin, SH},
  journal={IEEE Access},
  volume={11},
  pages={125094--125108},
  year={2023},
  publisher={IEEE}
}

@article{trauth2024reinforcement,
  title={A Reinforcement Learning-Boosted Motion Planning Framework: Comprehensive Generalization Performance in Autonomous Driving},
  author={Trauth, Rainer and Hobmeier, Alexander and Betz, Johannes},
  journal={arXiv preprint arXiv:2402.01465},
  year={2024}
}

@inproceedings{ding2020learning,
  title={Learning to collide: An adaptive safety-critical scenarios generating method},
  author={Ding, Wenhao and Chen, Baiming and Xu, Minjun and Zhao, Ding},
  booktitle={2020 IEEE/RSJ International Conference on Intelligent Robots and Systems (IROS)},
  pages={2243--2250},
  year={2020},
  organization={IEEE}
}

@inproceedings{pleines2023memory,
  title={Memory Gym: Partially Observable Challenges to Memory-Based Agents},
  author={Marco Pleines and Matthias Pallasch and Frank Zimmer and Mike Preuss},
  booktitle={International Conference on Learning Representations},
  year={2023},
  url={https://openreview.net/forum?id=jHc8dCx6DDr}
}

@inproceedings{krajewski2018highd,
  title={The highd dataset: A drone dataset of naturalistic vehicle trajectories on german highways for validation of highly automated driving systems},
  author={Krajewski, Robert and Bock, Julian and Kloeker, Laurent and Eckstein, Lutz},
  booktitle={2018 21st international conference on intelligent transportation systems (ITSC)},
  pages={2118--2125},
  year={2018},
  organization={IEEE}
}

@article{zhang2023recognition,
  title={Recognition method of abnormal driving behavior using the bidirectional gated recurrent unit and convolutional neural network},
  author={Zhang, Yu and He, Yingying and Zhang, Likai},
  journal={Physica A: Statistical Mechanics and its Applications},
  volume={609},
  pages={128317},
  year={2023},
  publisher={Elsevier}
}

@article{jing2024efficient,
  title={An Efficient High-Risk Lane-Changing Scenario Edge Cases Generation Method for Autonomous Vehicle Safety Testing},
  author={Jing, Shoucai and Zhao, Yuyu and Zhao, Xiangmo and Hui, Fei and Khattak, Asad J},
  journal={IEEE Transactions on Intelligent Vehicles},
  year={2024},
  publisher={IEEE}
}

@article{chen2021adversarial,
  title={Adversarial evaluation of autonomous vehicles in lane-change scenarios},
  author={Chen, Baiming and Chen, Xiang and Wu, Qiong and Li, Liang},
  journal={IEEE transactions on intelligent transportation systems},
  volume={23},
  number={8},
  pages={10333--10342},
  year={2021},
  publisher={IEEE}
}

@article{gan2022higan+,
  title={HiGAN+: handwriting imitation GAN with disentangled representations},
  author={Gan, Ji and Wang, Weiqiang and Leng, Jiaxu and Gao, Xinbo},
  journal={ACM Transactions on Graphics (TOG)},
  volume={42},
  number={1},
  pages={1--17},
  year={2022},
  publisher={ACM New York, NY}
}

@article{wang2021generative,
  title={Generative adversarial networks in computer vision: A survey and taxonomy},
  author={Wang, Zhengwei and She, Qi and Ward, Tomas E},
  journal={ACM Computing Surveys (CSUR)},
  volume={54},
  number={2},
  pages={1--38},
  year={2021},
  publisher={ACM New York, NY, USA}
}

@article{xu2021robust,
  title={Robust control of connected cruise vehicle platoon with uncertain human driving reaction time},
  author={Xu, Zhanrui and Jiao, Xiaohong},
  journal={IEEE Transactions on Intelligent Vehicles},
  volume={7},
  number={2},
  pages={368--376},
  year={2021},
  publisher={IEEE}
}

@article{stano2023model,
  title={Model predictive path tracking control for automated road vehicles: A review},
  author={Stano, Pietro and Montanaro, Umberto and Tavernini, Davide and Tufo, Manuela and Fiengo, Giovanni and Novella, Luigi and Sorniotti, Aldo},
  journal={Annual reviews in control},
  volume={55},
  pages={194--236},
  year={2023},
  publisher={Elsevier}
}

@article{van2022tractability,
  title={On the tractability of SHAP explanations},
  author={Van den Broeck, Guy and Lykov, Anton and Schleich, Maximilian and Suciu, Dan},
  journal={Journal of Artificial Intelligence Research},
  volume={74},
  pages={851--886},
  year={2022}
}

@article{team2023self,
  title={Self-Driving Car Technology for a Reliable Ride},
  author={Team, Waymo},
  year={2023}
}

@inproceedings{killing2021learning,
  title={Learning to robustly negotiate bi-directional lane usage in high-conflict driving scenarios},
  author={Killing, Christoph and Villaflor, Adam and Dolan, John M},
  booktitle={2021 IEEE International Conference on Robotics and Automation (ICRA)},
  pages={8090--8096},
  year={2021},
  organization={IEEE}
}

@article{ding2023survey,
  title={A survey on safety-critical driving scenario generation—A methodological perspective},
  author={Ding, Wenhao and Xu, Chejian and Arief, Mansur and Lin, Haohong and Li, Bo and Zhao, Ding},
  journal={IEEE Transactions on Intelligent Transportation Systems},
  volume={24},
  number={7},
  pages={6971--6988},
  year={2023},
  publisher={IEEE}
}

@article{zhao2023generalization,
  title={Generalization Generation of Hazardous Lane-changing Scenarios for  Automated Vehicle Testing},
  author={Zhao, xiaomo and Zhao, yuyu and Jing, shoucai and Huofei and Liu, jianbei},
  journal={ACTA AUTOMATICA SINICA},
  volume={49},
  number={10},
  pages={2211--2223},
  year={2023},
  publisher={ACTA AUTOMATICA SINICA}
}

@article{katariya2022deeptrack,
  title={Deeptrack: Lightweight deep learning for vehicle trajectory prediction in highways},
  author={Katariya, Vinit and Baharani, Mohammadreza and Morris, Nichole and Shoghli, Omidreza and Tabkhi, Hamed},
  journal={IEEE Transactions on Intelligent Transportation Systems},
  volume={23},
  number={10},
  pages={18927--18936},
  year={2022},
  publisher={IEEE}
}

@article{pronovost2023scenario,
  title={Scenario diffusion: Controllable driving scenario generation with diffusion},
  author={Pronovost, Ethan and Ganesina, Meghana Reddy and Hendy, Noureldin and Wang, Zeyu and Morales, Andres and Wang, Kai and Roy, Nick},
  journal={Advances in Neural Information Processing Systems},
  volume={36},
  pages={68873--68894},
  year={2023}
}

@phdthesis{heuer2022scenario,
  title={Scenario Generation for Testing of Automated Driving Functions based on Real Data},
  author={Heuer, Fin Malte},
  year={2022},
  school={Technische Universit{\"a}t Braunschweig}
}

@inproceedings{zhang2024chatscene,
  title={ChatScene: Knowledge-Enabled Safety-Critical Scenario Generation for Autonomous Vehicles},
  author={Zhang, Jiawei and Xu, Chejian and Li, Bo},
  booktitle={Proceedings of the IEEE/CVF Conference on Computer Vision and Pattern Recognition},
  pages={15459--15469},
  year={2024}
}

@inproceedings{song2022learning,
  title={Learning to predict vehicle trajectories with model-based planning},
  author={Song, Haoran and Luan, Di and Ding, Wenchao and Wang, Michael Y and Chen, Qifeng},
  booktitle={Conference on Robot Learning},
  pages={1035--1045},
  year={2022},
  organization={PMLR}
}

@article{lu2022vehicle,
  title={Vehicle trajectory prediction in connected environments via heterogeneous context-aware graph convolutional networks},
  author={Lu, Yuhuan and Wang, Wei and Hu, Xiping and Xu, Pengpeng and Zhou, Shengwei and Cai, Ming},
  journal={IEEE Transactions on intelligent transportation systems},
  volume={24},
  number={8},
  pages={8452--8464},
  year={2022},
  publisher={IEEE}
}

@article{zuo2023trajectory,
  title={Trajectory prediction network of autonomous vehicles with fusion of historical interactive features},
  author={Zuo, Zhiqiang and Wang, Xinyu and Guo, Songlin and Liu, Zhengxuan and Li, Zheng and Wang, Yijing},
  journal={IEEE Transactions on Intelligent Vehicles},
  year={2023},
  publisher={IEEE}
}

@article{hui2022deep,
  title={Deep encoder--decoder-NN: A deep learning-based autonomous vehicle trajectory prediction and correction model},
  author={Hui, Fei and Wei, Cheng and ShangGuan, Wei and Ando, Ryosuke and Fang, Shan},
  journal={Physica A: Statistical Mechanics and its Applications},
  volume={593},
  pages={126869},
  year={2022},
  publisher={Elsevier}
}

% biography section
% 
% If you have an EPS/PDF photo (graphicx package needed) extra braces are
% needed around the contents of the optional argument to biography to prevent
% the LaTeX parser from getting confused when it sees the complicated
% \includegraphics command within an optional argument. (You could create
% your own custom macro containing the \includegraphics command to make things
% simpler here.)
%\begin{IEEEbiography}[{\includegraphics[width=1in,height=1.25in,clip,keepaspectratio]{mshell}}]{Michael Shell}
% or if you just want to reserve a space for a photo:

% You can push biographies down or up by placing
% a \vfill before or after them. The appropriate
% use of \vfill depends on what kind of text is
% on the last page and whether or not the columns
% are being equalized.

%\vfill

% Can be used to pull up biographies so that the bottom of the last one
% is flush with the other column.
%\enlargethispage{-5in}

% that's all folks
\end{document}